%% file: main.tex
\documentclass[runningheads]{llncs}
\usepackage[utf8]{inputenc}
\usepackage{graphicx}
\usepackage{pgfplots} 
\pgfplotsset{width=4.5cm,compat=1.9}
\usepackage{pgfplotstable}
\usepgfplotslibrary{groupplots}
\usepackage{amsmath,amssymb}
\usepackage{multirow}
\usepackage[noadjust]{cite}
\usepackage{tikz}
\usepackage{hyperref}
\hypersetup{
	colorlinks,
	linkcolor={red!90!black},
	citecolor={green!80!black},
	urlcolor={blue!80!black}
}

\usepackage{makecell}
\usepackage{subcaption}

\newcommand{\red}{\color{red}}
\definecolor{pinegreen}{rgb}{0.0, 0.5, 0.0}
\newcommand{\green}{\color{pinegreen}}

\newcommand{\CF}{\operatorname{CF}}

\begin{document}

\title{On Sign Language Recognition from Skeletal Trajectory Data with Interpretable Classifiers}
\title{Exploring Sub-skeleton Trajectories for Interpretable Recognition of Sign Language}
% \subtitle{Do you have a subtitle?\\ If so, write it here}

\titlerunning{Exploring Sub-skeletons for Interpretable Recognition of Sign Language} % shorten heading
\author{Joachim Gudmundsson \and {Martin~P. Seybold} \and John Pfeifer}
\institute{University of Sydney, Australia\\\email{\{joachim.gudmundsson, mpseybold\}@gmail.com, johnapfeifer@yahoo.com}}

% \author{Joachim Gudmundsson}       
%  \affiliation{ \institution{School of Computer Science, University of Sydney}   \state{NSW} \country{Australia} }
%  \email{joachim.gudmundsson@sydney.edu.au} 

% \author{John Pfeifer}
%  \affiliation{ \institution{School of Computer Science, University of Sydney}   \state{NSW} \country{Australia} }
%  \email{johnapfeifer@yahoo.com}       

% \author{Martin P. Seybold}
%  \affiliation{ \institution{School of Computer Science, University of Sydney}   \state{NSW} \country{Australia} }
%  \email{martin.seybold@sydney.edu.au}  

% \institute{\at School of Computer Science, University of Sydney, Sydney, Australia
% \and
% J. Pfeifer \email{johnapfeifer@yahoo.com}}

\date{Received: date / Accepted: date}
% The correct dates will be entered by the editor

\maketitle
%%%%%%%%%%%%%%%%%%%%%%%%%%%%%%%%%%%%%%%%%%%%%%%%%%%%%%%%%%%%%%%%%%%%%%%%%%%%%%%%
\begin{abstract}
%%%%%%%%%%%%%%%%%%%%%%%%%%%%%%%%%%%%%%%%%%%%%%%%%%%%%%%%%%%%%%%%%%%%%%%%%%%%%%%%
Recent advances in tracking sensors and pose estimation software enable smart systems to use trajectories of skeleton joint locations for supervised learning.
We study the problem of accurately recognizing sign language words, which is key to narrowing the communication gap between hard and non-hard of hearing people.

Our method explores a geometric feature space that we call `sub-skeleton' aspects of movement.
We assess similarity of feature space trajectories using natural, speed invariant distance measures, which enables clear and insightful nearest neighbor classification.
The simplicity and scalability of our basic method allows for immediate application in different data domains with little to no parameter tuning.

We demonstrate the effectiveness of our basic method, and a boosted variation, with experiments on data from different application domains and tracking technologies.
Surprisingly, our simple methods improve sign recognition over recent, state-of-the-art approaches.

\end{abstract}

%%%%%%%%%%%%%%%%%%%%%%%%%%%%%%%%%%%%%%%%%%%%%%%%%%%%%%%%%%%%%%%%%%%%%%%%%%%%%%%%
\section{Introduction}\label{sec:intro}
%%%%%%%%%%%%%%%%%%%%%%%%%%%%%%%%%%%%%%%%%%%%%%%%%%%%%%%%%%%%%%%%%%%%%%%%%%%%%%%%
The problem of automatically and accurately identifying the meaning of human body movement has gained research interest due to advances in motion capture systems, artificial intelligence algorithms, and powerful hardware.
Wearable tracking technology, such as micro electromechanical systems, have been shrinking in size and cost while improving in accuracy and availability.
Less invasive motion capture systems such as Microsoft's Kinect or the Leap Motion controller have also been used to capture motion of human actors for recently released benchmark data sets (e.g. KinTrans \cite{kintrans2020}, LM  \cite{hernandez2020}, NTU RGB+D \cite{shahroudy2016}). 
The systems typically capture several input sources, e.g. RGB video and depth map, to track a bounding box of the actor and the skeletal joint positions therein, for every sensor frame.

This work studies the problem of recognizing patterns in human sign language.
Human sign languages are systems of communication that use manual movement patterns of arms and hands as well as non-manual elements, such as head, cheek and mouth posture, to convey meaning.
We focus on data sets that consist of extracted, single word labeled inputs along with the sequences of skeletal joint locations of the actors.
Our goals are to attain high classification accuracy with acceptable learning and query latency on diverse and evolving data sources, requiring little to no parameter tuning. 
In contrast to deep learning models that assume static environments,
we are particularly interested in simple, interpretable methods that, in turn, provide insight for curation of evolving, publicly available catalogs of sign language.

Though there are recent works~\cite{adaloglou2020, rastgoo2020} on sign language recognition from video data, our problem is often considered a special case of human action~\cite{presti2016} or gesture~\cite{mantecon2019} classification from \emph{skeleton-based} data, both receiving tremendous attention from the research community.
Improved solutions for this special case might provide the deaf community with computational tools that improve communication between signers and sign language illiterates~\cite{young2019, kushalnagar2018}.

\paragraph{The Core of the Problem}
% \begin{quote}
``One of the biggest challenges of using pose-based features is that semantically similar motions may not necessarily be numerically similar''~\cite{yao2011}.
% \end{quote}
This statement, that pervades action recognition articles in various forms, is a key motivation for our method.
Clearly, actors perform movement patterns with varying temporal windows and varying speed/pause modulations therein.
Hence (overly) simplistic similarity measures can easily be fooled to return meaningless values.

We approach this with \emph{speed-invariant} similarity measures that simultaneously capture the shape and spatial location of trajectories,  without the use of numeric frame numbers.

%%%%%%%%%%%%%%%%%%%%%%%%%%%%%%%%%%%%%%%%%%%%%%%%%%%%%%%%%%%%%%%%%%%%%%%%%%%%%%%%
\subsubsection{Related Work}   \label{sec:related-work}
%%%%%%%%%%%%%%%%%%%%%%%%%%%%%%%%%%%%%%%%%%%%%%%%%%%%%%%%%%%%%%%%%%%%%%%%%%%%%%%%
The survey by Presti and La Cascia~\cite[Table~$4$]{presti2016} provides an overview of human action recognition methods and their benchmark performance, and includes popular skeleton-based benchmark data sets such as Berkley's Multimodal Human Action Database (MHAD) and a contribution by the University of Central Florida (UCF)~\cite{ofli2013, ellis2013}.
It categorizes the known approaches into automatically
mined joint features~(M) as well as user-specified features on dynamics~(D) and on joints, where the latter is sub-categorized in Spatial Relations~(S), Geometric Relations~(G), and Key-Pose Dictionaries~(K).
We briefly discuss the leading benchmark methods, and closely related work.

`LDS-mSVM' (D)~\cite{chaudhry2013} considers trajectories as result
of Linear Dynamic Systems, on location and velocity variables that are sub-sampled from skeleton limbs over different time scales, on which Multiple Kernel Learning is used. 

% In automatically generating methods (M), ~\cite{ofli2014}
% Foremost method in (S)
`Riemann-$k$NN' (G)~\cite{devanne2014} fixes a skeleton-localized coordinate system, for transformation invariance.
Using differential geometry, spatial trajectories are considered as curves in the associated velocity space that, after speed vector scaling, all have unit $L_2$ function norm.
$k$NN classification is then applied to the resulting velocity curves, using the `elastic shape' distance~\cite[Section~III]{devanne2014}.
% Then velocity curves are assessed by the `elastic shape' distance
% %{\gray , computed by finding an optimal orientation-preserving diffeomorphism with dynamic programming,} 
% in the $k$NN classification~\cite[Section~III]{devanne2014}.

Category (S) includes recent methods using Graph Convolutional Networks which were successfully adapted to architectures with one vertex per joint and frame.
The `ST-GCN' architecture~\cite{yan2018} contains so-called spatial edges, between neighboring skeleton joints, and temporal edges, between a joint's previous and next frame neighbor.
The spatial and temporal convolutions are interleaved, and long-range spatial edges have been introduced independently by~\cite{li2018}.
% 2019 works on ST-GCN
% AS-GCN: adjacency powering + pose dictionary
% STGR: additional edges `frame-wise attention and global self-attention'
% 2s-AGCN: 
% DGNN,
% GR-GCN
% 2020 works in ST-GCN
% MS-G3D  introduces long range cross-space-time edges
Improvements of this method were achieved based on 
key poses, 
frame-wise and sequence-wise attention edges, 
adaptive self-attention, 
motifs,
bone features, 
{$3$-frame} temporal look-ahead, 
Neural Searching,
and cross space-time edges spanning up to a user-specified hop distance~\cite{li2019-stgr, li2019actional, shi2019-dgnn, wen2019graph, shi2019,  gao2019-gr-gcn, peng2020learning, liu2020}.
Recently \cite{hernandez2020} provided `Kinematics-LSTM' (S) that, using convolutional LSTM networks on kinematic features, achieves highly accurate sign language recognition from Leap Motion skeleton data.

%%%%%%%%%%%%%%%%%%%%%%%%%%%%%%%%%%%%%%%%%%%%%%%%%%%%%%%%%%%%%%%%%%%%%%%%%%%%%%%%
\subsubsection{Contribution and Paper Structure}
%%%%%%%%%%%%%%%%%%%%%%%%%%%%%%%%%%%%%%%%%%%%%%%%%%%%%%%%%%%%%%%%%%%%%%%%%%%%%%%%
\begin{figure}[t]
    \centering
    \includegraphics[width=0.98\columnwidth]{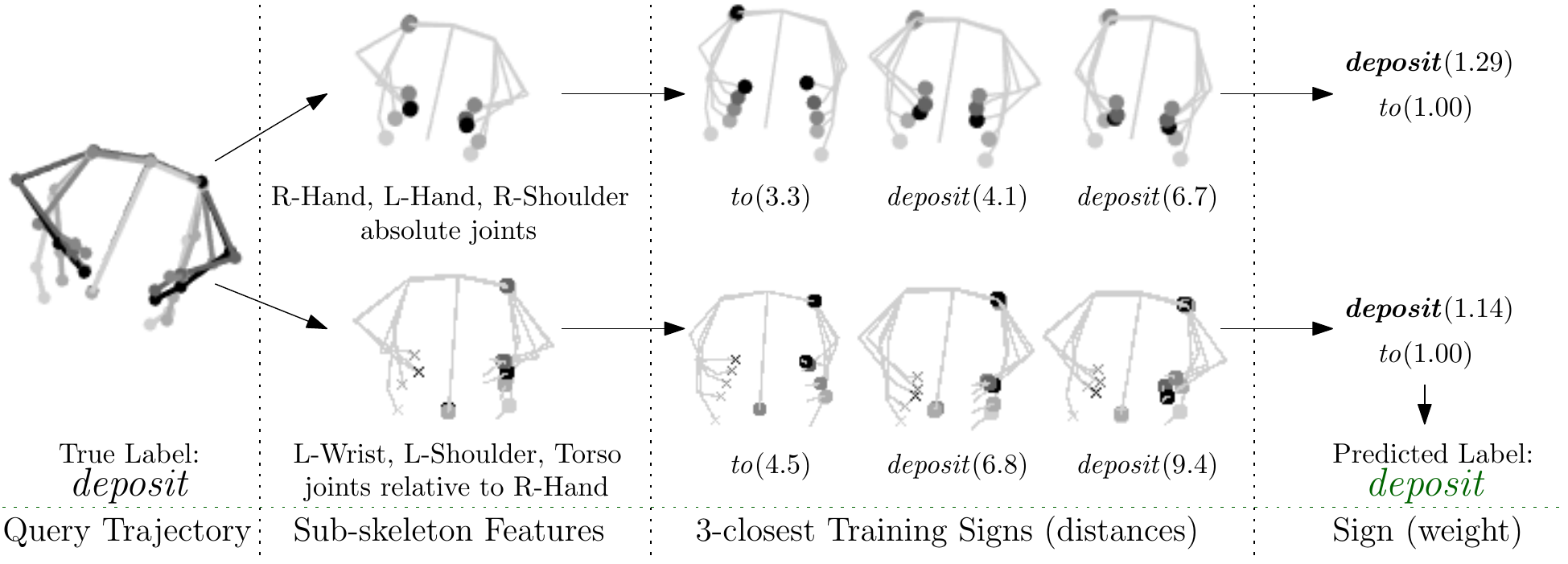}
    \\ \vspace{-.3cm}
    \caption{Example of the proposed $k$NN-m classification ($k=3$) of a query trajectory based on speed-invariant distances of one \emph{absolute} (top) and one \emph{relative} (bottom) sub-skeleton movement pattern (cf. Section~\ref{sec:classification} and Table~\ref{tab:interpretability_eg}).}
    \label{fig:overview_approach}
\end{figure}

The main idea of our method is to map human body skeletal movement into a feature space that captures absolute and relative movement of sets of joints (see Figure~\ref{fig:overview_approach}).
Based on the high-dimensional feature trajectories, we apply simple classifiers that use geometric similarity measures (see Section~\ref{sec:distances}).
The main technical difficulty is determining the most discriminative way to transform spatial joint locations into high-dimensional trajectories.
Motivated by sign language, we effectively navigate the vast space of absolute and relative movement patterns that subsets of joints describe, which we call `sub-skeleton' features.

We propose a general, novel method that \emph{automatically mines} a set of maximal discriminating sub-skeleton features, irrespective of the present skeleton formats
(see Section~\ref{sec:feat_space}). 
Our main contributions are as follows:

\begin{itemize}
\item 
Our mining method discovers sub-skeleton aspects in the data sets that are highly discriminating.
Since all feature trajectories directly relate to the input data, the merits of each classification result can be \emph{interpreted} and visualized naturally in terms of a geometric similarity measure (e.g. Figure~\ref{fig:example_distances}).%
\item 
Simple Nearest Neighbor and Ensemble Boosted Classification (see Section~\ref{sec:classification}) achieve improved accuracy on sign-language benchmarks of diverse tracking technologies. 
Competitive accuracy on Human Action benchmarks show that our method generalizes to other recognition problems.
Particularly noteworthy are our high accuracy results on \emph{very small} training sets (see Section~\ref{sec:exp_res}).

\item
To the best of our knowledge, we are the first to assess sub-skeleton features with trajectory similarity measures (e.g. the Fréchet distance) for sign language and human action classification problems.

\end{itemize}

Furthermore, our publicly available\footnote{\url{https://github.com/japfeifer/frechet-queries}} implementation achieves average query latency below $100$ms on standard, non-GPU, desktop hardware.

%%%%%%%%%%%%%%%%%%%%%%%%%%%%%%%%%%%%%%%%%%%%%%%%%%%%%%%%%%%%%%%%%%%%%%%%%%%%%%%%
\section{Setup and Problem Definition} \label{sec:distances}
%%%%%%%%%%%%%%%%%%%%%%%%%%%%%%%%%%%%%%%%%%%%%%%%%%%%%%%%%%%%%%%%%%%%%%%%%%%%%%%%
A skeleton $G$ is an undirected connected graph where each vertex represents a joint.
In our setting $G$ models a part of a human body skeleton where vertices are adjacent if their joints have a rigid connection (e.g. a bone).
As input for a word signed by an actor, we are given a \emph{sequence} $S$ of $n$ frames $\langle G_1, \ldots , G_n\rangle$, and each $G_i$ holds the $3$D location of the joints at time step $i$.
Frame frequency is typically constant but depends heavily on the capture technology.
Sequences have varying duration, for example, between $0.2$ and $30$ seconds (cf. Table~\ref{tab:datasets}).
Every sequence is labeled with one class and the actor's body movement is performed once, or a small number of times, per sign class.

The input data $\mathbb{D}$ is a set of sequences $\{S_1, \ldots , S_{\ell}\}$ and is partitioned into two sets: the training set $\mathbb{D}_r$ and the testing set $\mathbb{D}_t$.
Our aim is to classify each sequence in the test set, using only sequences from the training set, as accurately and efficiently as possible.

\subsubsection{Similarity Measures of Trajectories}
\begin{figure} \centering
    \includegraphics[width=\columnwidth]{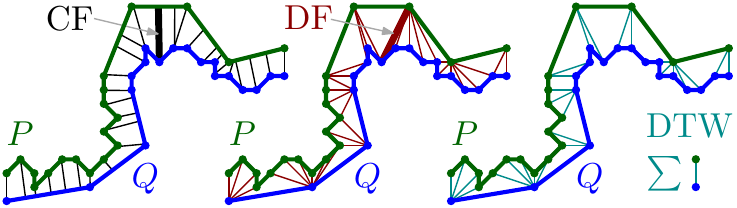}
    \\ \vspace{-.3cm}
    \caption{
    Comparison of CF, DF, and DTW measures on two visually similar trajectories $P$ (moves slow then fast) and $Q$ (moves fast then slow) with a total of $19$ and $18$ frames, respectively.
    }   \label{fig:example_distances}
\end{figure}

Various motion capture devices differ when comparing their frame rates and movement capture accuracy. 
Furthermore, some actors perform motions faster than others for the same class, even though they describe the same spatial pattern. Our proposed method converts every input sequence of frames into one or more feature trajectories.
Therefore, we are interested in trajectory similarity measures that are speed-invariant.
The continuous Fréchet~\cite{alt1995} distance
% and continuous Dynamic Time Warping~\cite{munich1999-cdtw} 
provides this property in a strict sense. 
We do, however, also discuss simpler discrete Fréchet and Dynamic Time Warping measures due to their popularity and suitability, especially when compared to other methods such as computing equally sampled Euclidean distances over a sequence.
Below we discuss three trajectory distance measures that we tested with our method: continuous Fréchet (CF) distance, discrete Fréchet (DF) distance and Dynamic Time Warping (DTW) (e.g. Figure~\ref{fig:example_distances}).
The first two are metrics, while the latter is not.

A $d$ dimensional trajectory $P$ of size $n$ is a directed polygonal curve through a sequence of $n$ vertices (points) $\langle p_1, \ldots , p_n\rangle$ in $\mathbb{R}^d$ and each contiguous pair of vertices in $P$ is connected by a straight line segment. 

\paragraph{Continuous Fréchet (CF) Metric.}
The continuous Fréchet distance between two trajectories $P,Q$ in $\mathbb{R}^{d}$ can be intuitively explained as follows.
Imagine one person walks along $P$ and another person walks along $Q$.
Each person must walk along his or her curve from start to finish.
Neither person is allowed to 
stand still or 
travel backwards, but otherwise they are free to vary their speed.
The cost of any fixed walk is the maximum distance that is attained between the two people at any time during the walk.
Different walks can have different costs, and the continuous Fréchet distance between $P$ and $Q$, denoted $\CF(P,Q)$, equals the minimum possible cost over all walks.
More formally, we have 

\begin{equation*}
 \CF(P,Q)=\inf_{h}\max_{p\in P} \Big\lVert p, h(p) \Big\rVert,
\end{equation*}

where $\lVert\cdot,\cdot\rVert$ denotes the Euclidean distance and $h:P\rightarrow Q$
is a homeomorphism that assigns to every point $p \in P$
a point $h(q)\in Q$.

The continuous Fréchet distance is a (pseudo) metric and can be computed in $\mathcal{O}(mn \log mn)$ time, where the $m$ and $n$ are the sizes of the two trajectories. 

\paragraph{Discrete Fréchet (DF) Metric.}
This distance~\cite{eiter1994} is very similar to CF, except that the two persons are jumping along their trajectories from vertex to vertex, may stand still at a vertex, and are never considered to be in the interior of a line segment.
It can be computed in $\mathcal{O}(mn)$ time.

\paragraph{Dynamic Time Warping (DTW).} 
The DTW similarity measure~\cite{keogh2005} only differs slightly from the discrete Fréchet distance in replacing the operation `maximum' with ‘summation’ for the distances of the matched vertices. As a result DTW does not fulfill the triangle inequality, and is therefore not a metric. It can be computed in $\mathcal{O}(mn)$ time.

\paragraph{Properties of the trajectory distance measures.}
The three distances are designed to measure the spatial similarity between trajectories and do not take time into consideration.  
The different measures have various advantages and drawbacks.
The main advantage with the DTW measure is that it is less impacted by outliers and ``smooths over'' significant differences in comparison to the Fréchet distances which capture the worst-case changes between two trajectories.
However, DTW is not a metric, and two visually similar trajectories may result in a large measure if the frame rate varies or the same action is performed tremendously faster by one subject than by another.

CF and DF are metrics and can take advantage of metric indexing~\cite{gudmundsson2020} which provides very fast query times for $k$-Nearest~Neighbor computations. The CF distance is generally regarded to capture visual similarity more precisely than DF, especially if the sampling rate is very sparse, since it takes the continuity of the shapes into account.

% \clearpage
%%%%%%%%%%%%%%%%%%%%%%%%%%%%%%%%%%%%%%%%%%%%%%%%%%%%%%%%%%%%%%%%%%%%%%%%%%%%%%%%
    \section{Mining Sub-Skeleton Features} \label{sec:feat_space}
%%%%%%%%%%%%%%%%%%%%%%%%%%%%%%%%%%%%%%%%%%%%%%%%%%%%%%%%%%%%%%%%%%%%%%%%%%%%%%%%

Sign language conveys meaning based on the movement patterns of skeletal joints. 
Our goal is to mine joint movement and determine which combination of joints best discriminates classes. 
We first describe three feature mining concepts that are important in our setting, and then present our mining algorithm.

\paragraph{Absolute and Relative Joints.}
The movement of a joint can be viewed in an absolute space or relative space.
We say that a joint is absolute if its position is described in a fixed coordinate system.
For example, the left index fingertip joint moving through space is an absolute joint.
A joint whose movement is described in terms of its position relative to another joint whose spatial location serves as the center of a moving coordinate system is called a relative joint, e.g., the joints of the right arm relative to the neck joint.

\paragraph{Feature Space.}
Consider an underlying feature space $F$ that contains all possible absolute and relative joint subset combinations derived from skeleton $G$.
Note that $F$ contains $|G|^2$ \emph{singletons}, i.e., $|G|$ absolute and $|G|^2 - |G|$ relative joints.
Any \emph{feature} $f \in F$ can be derived as the union of $|f|$ singletons, regardless of $f$ being an absolute or relative sub-skeleton.
Thus, the size of $F$ is \emph{exponential} in the number of joints $|G|$ of the skeleton.

\paragraph{Feature Trajectory.}
Any given feature $f \in F$ maps input skeletal movement to feature trajectories, i.e. $f$ maps $\langle G_1,\ldots,G_n\rangle$ to $\langle p_1,\ldots,p_n\rangle$ and each point $p_i$ has dimensionality $3|f|$, where $|f|$ denotes the number of joints in the set.

\subsubsection{Greedy Feature Template Mining Algorithm}
\label{ssec:sl_feat_temp}
Since the feature space is huge, we desire an algorithm that quickly chooses a small number of features from $F$, and achieves high accuracy on a classifier that is simple and interpretable.
Thus, we now describe our mining algorithm that \emph{greedily} selects a small number of discriminating features $\{f_1,\ldots,f_l\}$ from $F$.

The set of \emph{canonical sub-skeletons} $C \subseteq F$ form the basis of our greedy exploration.
In the case where $|G|$ is not too large, $C$ contains all singletons.
In the case where $|G|$ is very large, we derive a smaller set $C$ by merging similar singletons together into a single set (e.g., four singleton joints of the index finger are merged into a single `index finger' set).
The singleton groups are computed by identifying all chains of degree $2$ vertices in $G$ (e.g., right leg, left arm), and for the case of deriving relative canonical sub-skeletons we use central joints (e.g., neck or torso) as reference joints.
Figure~\ref{fig:feature_space} shows an example of merging similar hand singletons together to reduce the size of $C$.

The mining algorithm computes a discriminating \emph{feature template} $T\subseteq F$.
There are up to $1+|G|$ features in $T$: one feature that contains the union of one or more \emph{absolute} singletons, and $|G|$ \emph{relative} features, each of which contain the union of one or more singletons that have the \emph{same} reference joint.
We denote with $l(T)$ the number of features from $F$ that are defined by $T$, so $l(T) \leq 1+|G|$.

The algorithm constructs $T$ by iteratively performing \emph{adapted union} operations, denoted $\Tilde{\cup}$.
In this context, a given relative joint $c \in C$ is added to the feature $f \in T$ that has the same reference joint, and a given absolute joint $c \in C$ is added to the feature $f \in T$ that contains the absolute joints.
For example, say $T$ contains two features $f_1$ and $f_2$, where $f_1 \in T$ is relative to the right elbow and $f_2 \in T$ contains absolute joints.
If a joint $c$ is relative to the right elbow, then it will be added to $f_1$.
If a joint $c$ is an absolute joint, then it will be added to $f_2$.
If a joint $c$ is relative to the neck, then it will be added to an initially empty feature $f_3$.

\input{figures/feature_space}

We use standard classifiers to determine which elements of $C$ to add to the feature template.
The simplest is a Nearest~Neighbor method that finds the closest label in $\mathbb{D}_r$ based on a trajectory distance measure (see Section~\ref{sec:classification} for the precise description of our classifiers).
To employ underlying classifiers during our greedy exploration, we initially partition the training set $\mathbb{D}_r$ randomly with a $1$:$2$-split into $\mathbb{D}'_r$ and $\mathbb{D}'_t$
and proceed as follows:

\begin{enumerate}
    \item Start with the empty set $T=\emptyset$.
    \item Compute $\forall$ $c \in C\setminus T$ the classification accuracy of $T $ $\Tilde{\cup}$ $c$ on $\mathbb{D}_r'$ and $\mathbb{D}_t'$.  \label{algo:body}
    \item If one such best $c$ improves over the last iteration, \\ then add $c$ to $T$ and GOTO~\ref{algo:body}.
    \item Return $T$.
\end{enumerate}

The resulting set $T$, which contains a subset of canonical sub-skeletons, describes a small number of absolute and relative features by means of the aforementioned adapted union operation.
Hence we have an equal number of generated feature trajectories for each input sequence.
Clearly, the algorithm can be executed for different classifiers and trajectory similarity measures to select the best feature set found.

Since this deterministic greedy exploration is based on very simple classifiers, with very few or no parameters, the `overfitting' problem of complex models is avoided.
Nevertheless, the obtained classification accuracy is already comparatively high (see Section~\ref{sec:exp-accuracy}).

Note that the construction of the feature template from $\mathbb{D}'_r$ and $\mathbb{D}'_t$ is only to accommodate a fair comparison of the final accuracy against other recognition methods.
For the purpose of pure pattern discovery, one may well use the whole data set $\mathbb{D}$ in the selection process.

% \clearpage
%%%%%%%%%%%%%%%%%%%%%%%%%%%%%%%%%%%%%%%%%%%%%%%%%%%%%%%%%%%%%%%%%%%%%%%%%%%%%%%%
    \section{Classifying Feature Trajectories} \label{sec:classification}
%%%%%%%%%%%%%%%%%%%%%%%%%%%%%%%%%%%%%%%%%%%%%%%%%%%%%%%%%%%%%%%%%%%%%%%%%%%%%%%%
We focus on classifiers that are solely based on geometric similarity measures between the feature trajectories of the query and training sequences.
This not only provides results that can be interpreted and visualized, but also allows us to employ clustering-based metric indexing techniques to achieve very small query latency (e.g. Section~\ref{sec:query-time}).
For these, we describe simple $k$-Nearest~Neighbor and ensemble boosted classification techniques.
Both may achieve slightly improved accuracy by optional pre-processing with natural normalizations, which we describe next.

The spatial coordinates of a sequence $S$ are normalized with a translation, rotation, standardizing the limb lengths, and padding with an at-rest pose. 
Each is independently applicable.
Variations of the first three are common~\cite{presti2016}, hence we only discuss our at-rest pose briefly. 
Two sequences for the same class may have similar motion in the middle of their sequence, but different start or end frame joint locations. 
This can result in a large trajectory distance even though they are the same signed word. 
An at-rest pose is a standard skeletal pose frame that depicts the subject in a relaxed or neutral state, and can result in a smaller trajectory distance. 
Each data set has one at-rest pose definition frame, which is inserted at the beginning and end of each sequence. 

Our {\bf $k$-Nearest~Neighbor single} ($k$NN-s) classifier simply concatenates the spatial coordinates of the $l(T)$ feature trajectories frame-wise to derive a \textit{single} trajectory for each input sequence. 
Then, for a given query sequence, we use a trajectory similarity measure to retrieve the $k$-closest sequences from the training set and return the label with the maximum weight, computed as follows.
Determine the smallest distance $\bar{d}$ from the $k$ values.
For each encountered label compute the weight $\sum_i \bar{d}/d_i$, where the sum is over the distance values $d_i$ of entities with this label.
Our {\bf $k$-Nearest~Neighbor multi} ($k$NN-m) classifier retrieves the $k$-closest sequences from the training set for each of the $l(T)$ features separately.
Then we predict one label for each the $l(T)$ features with the approach above and use a majority vote over the results to obtain the final label.
%(ties are broken by choosing the predicted label containing the smallest distance).
See Table~\ref{tab:interpretability_eg} for $k$NN-m examples which demonstrate that the classification is transparent and interpretable.

\input{tables/interpretability_eg.tex}

Our {\bf Distance Matrix} (DM) \textbf{single} and \textbf{multi} methods exploit computed trajectory similarity values, between query sequence and training sequences, using more advanced, multiclass classifiers (e.g. SVM or the random subspace linear discriminant classifier) to boost accuracy results.
For DM-s we use aforementioned concatenation, to assemble a $|\mathbb{D}_r|\times|\mathbb{D}_r|$ training matrix of geometric similarity values, 
which is the basis of the multiclass learning.
For a classification query, we compute the row-vector of geometric distance values and then apply the learned classifier to retrieve the final label.
Our DM-m method considers the similarity values to the individual feature trajectories based on the $|\mathbb{D}_r|\times l(T)|\mathbb{D}_r|$ matrix used for the multiclass learning\footnote{Our experiments use Matlab's \texttt{fitcensemble}, except for NTU60 which uses SVM.}.

The classification methods (from $k$NN-s, to $k$NN-m, to DM-s, to DM-m) contain an increasing amount of information, respectively, with a cost of more distance computations and classifier overhead (see Figure~\ref{fig:testtime_small} for query latency).

%%%%%%%%%%%%%%%%%%%%%%%%%%%%%%%%%%%%%%%%%%%%%%%%%%%%%%%%%%%%%%%%%%%%%%%%%%%%%%%%
    \section{Experimental Setup and Results} \label{sec:exp_res}
%%%%%%%%%%%%%%%%%%%%%%%%%%%%%%%%%%%%%%%%%%%%%%%%%%%%%%%%%%%%%%%%%%%%%%%%%%%%%%%%

We implemented our methods and ran experiments on five skeleton-based data sets of actors that perform American Sign Language and Human Actions, each captured with different tracking technologies.
Through experiments, we investigate if
(i)  sub-skeleton feature mining discovers highly discriminative movement patterns, 
(ii) classification accuracy improves on state-of-the art methods for the benchmarks and on small training sets,
and (iii) classification queries are answered quickly.

Experiments ran on a standard desktop computer ($3.6$GHz Core i7-7700 CPU, non-GPU) with MATLAB R$2020$a software on Windows 10,
except for those using competitor ST-GCN~\cite{yan2018}, which ran on a high-performance computing environment with an NVIDIA V100 GPU and up to 32 CPUs (we attempted to run the ST-GCN author's code on the desktop computer but the learning did not converge within three days).

%%%%%%%%%%%%%%%%%%%%%%%%%%%%%%%%%%%%%%%%%%%%%%%%%%%%%%%%%%%%%%%%%%%%%%%%%%%%%%%%
    \subsubsection{Data Sets} \label{sec:exp_res:datasets}
%%%%%%%%%%%%%%%%%%%%%%%%%%%%%%%%%%%%%%%%%%%%%%%%%%%%%%%%%%%%%%%%%%%%%%%%%%%%%%%%
We use the skeleton joint-based Human Sign Language and Human Action data sets introduced in Section~\ref{sec:intro}, see Table~\ref{tab:datasets} for an overview.

\input{tables/datasets.tex}

The American sign language data set KinTrans is exclusively obtained from \cite{kintrans2020}, hence we are the first to publish results on it.
Skeletal movement is captured with a Microsoft Azure Kinect DK by sign-literate individuals in a controlled setting. 
Some of the sequences contain noise and subject IDs are not available due to privacy.
Only $10$ joints are captured: right/left hand/wrist/elbow/ shoulder, neck and hip center. 
Recognizing signed words can be difficult since many signs use detailed finger movement, which is not captured with this data set.
The American sign language data set LM~\cite{hernandez2020} captures detailed right/left finger, wrist, and elbow joint movement using the Leap Motion controller. 
A data sequence contains one subject repeating a particular sign a variable number of times, and are `cut' into segments that contain the single sign just once.
Hernandez et al.~\cite{hernandez2020} use a manually created segmentation, which is not publicly available.
Therefore, we use the following automated segmentation.
Transform a raw sequence into a $1$D signal by converting each frame into a $3|G|$ dimensional point which is used to compute the Euclidean distance to the first point of the sequence.
Then use the mode in the Fourier power spectrum to obtain the number of segments that are aligned greedily to capture local minima of the $1$D signal
(e.g. Figure~\ref{fig:LM_cut_eg}).
Our automated cuts yielded slightly more sequences (17,312) than the LM author's manual cut (16,890).
\input{figures/LM_cut_eg}

The NTU RGB+D human action data set~\cite{shahroudy2016} captures subject movement using Microsoft Kinect v2 sensors.
An action sequence is captured simultaneously by three camera views: one directly facing the subject, and the other two at $45$ degree angles.
There are $17$ different camera setups, single and multi-subject action classes, and subject ages range from $10$ to $35$ years old.
We study single-subject sequences, therefore NTU RGB+D sequences with two or more detected subjects are removed, and we name the resulting data set NTU60.
We also test on popular human action data sets from the survey~\cite{presti2016}.
The University of Central Florida~(UCF) data set~\cite{ellis2013} captures movement with a Kinect sensor, at $30$Hz, % and estimates skeletal joints with 
using OpenNI, and Berkeley's Multimodal Human Action Database~(MHAD) captures motion at $480$Hz with five different systems~\cite{ofli2013}.  

To speed up distance computations, we use trajectory simplifications for LM, UCF, and MHAD.
Simplification reduces the size of trajectories at the cost of a small error.
We use the algorithm of~\cite{agarwal2005} as it has a guaranteed bounded error under the continuous Fréchet distance. 

%%%%%%%%%%%%%%%%%%%%%%%%%%%%%%%%%%%%%%%%%%%%%%%%%%%%%%%%%%%%%%%%%%%%%%%%%%%%%%%%
    \subsubsection{Feature Template Mining}     \label{sec:exp_res:mining}
%%%%%%%%%%%%%%%%%%%%%%%%%%%%%%%%%%%%%%%%%%%%%%%%%%%%%%%%%%%%%%%%%%%%%%%%%%%%%%%%

{For $k$NN-s,}
the template mining using DTW provided more accurate results for KinTrans, NTU60 and UCF data sets whereas the CF distance was better for LM and MHAD data sets.
KinTrans is noisier than LM which may account for the different choice of similarity measures between these two data sets.
For our DM classification methods, the SVM and subspace discriminant classifier was more accurate for the NTU60 and UCF data set, respectively, which shows that our mining process can select different DM classifiers depending on the underlying data.
Interestingly, normalizations are completely discarded on two data sets and none use all normalizations.
Hence, our feature template mining can remove the need for data normalizations. 

\input{tables/feature_mining_results.tex}

The number of absolute and relative joints in selected feature templates are small compared to $|C|$, which results in lower dimensional trajectory features and faster distance computations (see Table~\ref{tab:feature_mining_results}).
The most discriminative KinTrans joint is the right hand and for LM it is the right index finger, but interestingly, KinTrans neck and torso joints were also included (which only have subtle movement), and the UCF and MHAD feature templates contain mostly different joints compared to each other.

The KinTrans feature mining took $47$ minutes for a specific normalization combination, which we consider reasonable on a standard desktop computer.
As a guide, the respective ST-GCN training in Table~\ref{tab:accuracy_sign_lang} took more than $11$ hours on the high-performance GPU computing environment.

% \clearpage
%%%%%%%%%%%%%%%%%%%%%%%%%%%%%%%%%%%%%%%%%%%%%%%%%%%%%%%%%%%%%%%%%%%%%%%%%%%%%%%%
    \subsubsection{Classification Accuracy}\label{sec:exp-accuracy}
%%%%%%%%%%%%%%%%%%%%%%%%%%%%%%%%%%%%%%%%%%%%%%%%%%%%%%%%%%%%%%%%%%%%%%%%%%%%%%%%

% \input{tables/accuracy_lm.tex}
\input{tables/accuracy_sign_lang.tex}
We compare the accuracy of our simple classifiers against various state-of-the-art skeleton-based recognition approaches discussed in Section~\ref{sec:related-work}, using five diverse data sets from Table~\ref{tab:datasets}.
Experiments on the new KinTrans data set gradually increase the amount of training data for each subsequent validation protocol, i.e. randomly chosen training sets that contain either $2, 3, 10\%$, or $20\%$ sequences per class and use the remaining sequences as the respective testing sets.
Experiments with previously released data sets use \emph{exactly} the same validation protocols as in the comparison papers: $5$-fold cross-subject for LM, cross-subject and cross-camera view for NTU60, $4$-fold cross for UCF, and first seven subjects for training for MHAD.
In all DM-m classifier experiments, we use the full number of distance matrix columns, except for LM where we reduced it by randomly choosing two sequences per class/subject.

To investigate the effectiveness of our method on sign language data sets, we compare our accuracy results against: 
(i) publicly available code of the recent ST-GCN method (that exclusively supports body skeletons) for our KinTrans data set, and (ii) LM data set accuracy results of~\cite{hernandez2020} (see Table~\ref{tab:accuracy_sign_lang}).
The KinTrans experiment investigates the effectiveness of our method on varying training set sizes compared to ST-GCN. 
Training and testing data for each validation protocol experiment are fed into the ST-GCN application environment in NTU skeletal format, using all KinTrans joints without information loss.
To achieve the best possible ST-GCN results, we tuned the learning parameters, replicated each sequence in the training sets exactly $40$ times, and report the best accuracy result out of three independent training runs per experiment.

In the lowest training information experiment ($2$ trainers per class) we have almost $40\%$ accuracy improvement over ST-GCN.
As more training information is introduced, we still outperform ST-GCN, even with our simple $k$NN-s classifier.
Experiment results show that simpler $k$NN-s classifiers can often achieve similar accuracy results compared to the more complex DM-m method, and in some cases the $k$NN-m outperforms the $k$NN-s under the CF distance.
For the LM sign language data set, our DM-m method improves on the accuracy results of~\cite{hernandez2020} which uses a more complex LSTM neural network model.

Table~\ref{tab:accuracy_human_action} compares our accuracy results against three human action data sets.
We achieve $99.5\%$ and $100\%$ accuracy on the UCF and MHAD data sets, respectively, which slightly improves on published results (i.e. values in italics).
For the NTU60 data set we achieve accuracy results that are similar to the performance of ST-GCN, though this benchmark contains sequences that perform a variable number of repetitions of the action and our methods were only designed for sign language.

The experiments show that the accuracy performance of our simple methods generalizes well over different data domains (sign language/human action), training set sizes (small/large), and skeleton capture formats (coarse body/detailed hands).

%%%%%%%%%%%%%%%%%%%%%%%%%%%%%%%%%%%%%%%%%%%%%%%%%%%%%%%%%%%%%%%%%%%%%%%%%%%%%%%%
\subsubsection{Query Time}\label{sec:query-time}
%%%%%%%%%%%%%%%%%%%%%%%%%%%%%%%%%%%%%%%%%%%%%%%%%%%%%%%%%%%%%%%%%%%%%%%%%%%%%%%%

\input{figures/testtime_small.tex}

To investigate the query latency of our distance based classifiers, we run experiments on training sets of different sizes from the KinTrans data set with each of our methods.
The testing sets always contain the remainder of the whole data set and the DM methods compute the full number of distance columns.
For the metric distance measure CF, we use the $k$~Nearest-Neighbor search structure from~\cite{gudmundsson2020}.
The results in Figure~\ref{fig:testtime_small} show average wall-clock time per query, overall classification accuracy, and average number of necessary distance computations per query for our 1NN-s (blue) and DM methods (black). 
All 1NN-s and DM-s classifiers show an average query time under $200$ms.

%%%%%%%%%%%%%%%%%%%%%%%%%%%%%%%%%%%%%%%%%%%%%%%%%%%%%%%%%%%%%%%%%%%%%%%%%%%%%%%%

% \clearpage
%%%%%%%%%%%%%%%%%%%%%%%%%%%%%%%%%%%%%%%%%%%%%%%%%%%%%%%%%%%%%%%%%%%%%%%%%%%%%%%%
\section{Conclusion and Future Work}
%%%%%%%%%%%%%%%%%%%%%%%%%%%%%%%%%%%%%%%%%%%%%%%%%%%%%%%%%%%%%%%%%%%%%%%%%%%%%%%%

Our work on skeleton-based sign language recognition
introduced the sub-skeleton feature space and studied it using speed-invariant similarity measures.
Our method automatically discovers absolute and relative movement patterns, which enables highly accurate Nearest~Neighbor classification, with acceptable latency, on training data of varying domains, skeleton formats, and sizes. 
Our distance based classifiers are interpretable and train on basic computing hardware, which make them particularly interesting for data sets that change frequently.

In future work we seek to extend our methods to recognition problems that do not assume segmented input data.
Popular skeleton-based benchmark data has variable repetitions of the action in each recording, e.g. NTU120 and Kinetics~\cite{kay2017kinetics,liu2019ntu}, or contains continuous streams of different sign language words per recording~\cite{pu2018dilated,wei2019deep}.
To accommodate such recognition problems, we seek to extend our automated, Fourier based, segmentation method or investigate sub-trajectory clustering approaches.

\subsubsection{Acknowledgements}
The authors acknowledge the technical assistance provided by the Sydney Informatics Hub, a Core Research Facility of the University of Sydney.
This work was supported under the Australian Research Council Discovery Projects funding scheme (project number DP180102870).

\vspace{-.5cm}
\bibliographystyle{splncs04}
\bibliography{references}

\clearpage
\appendix
\input{sec-supplement.tex}

\end{document}

%% file: figures/feature_space.tex
\begin{figure*}[t]\centering
    % \includegraphics[width=\textwidth % ,clip,trim=300 0 0 0 
    % ]{./figures/pic1.png} \\
    \includegraphics[width=\textwidth]{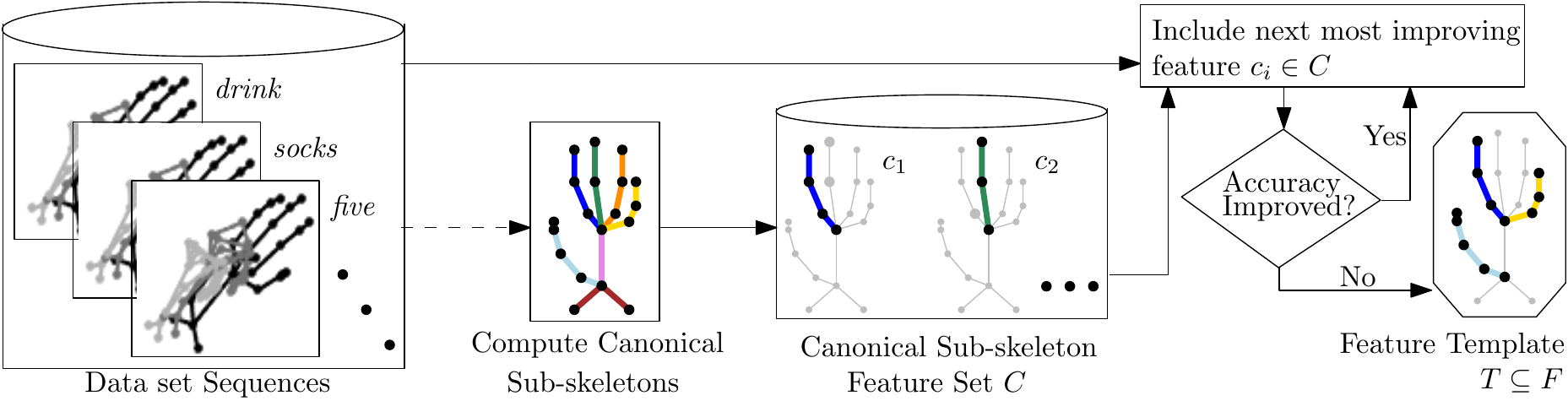}
\\ \vspace{-.3cm}
\caption{
Automated mining from sub-skeleton features produces a highly discriminative Feature Template.
A feature set $C$ of canonical sub-skeletons is determined from skeletal joints from $G$.
Then, a greedy algorithm constructs a feature template $T$ by testing and adding the next best feature $c_i \in C$, until the accuracy no longer improves.
}

	\label{fig:feature_space} 
\end{figure*}

% \begin{figure}[t]\centering
%     % \includegraphics[width=\textwidth % ,clip,trim=300 0 0 0 
%     % ]{./figures/pic1.png} \\
%     \includegraphics[width=\textwidth]{figs/feat_space_template.pdf}

% \label{fig:feature_space} 
% \end{figure}

%% file: tables/interpretability_eg.tex
\begin{table}[t]
  \centering
{
\begin{tabular}{ll|c|c}
\hline
~ & True      & \multicolumn{2}{c}{$k$-closest sequence labels and distances}                      \\
  & Label     & R-Hand, L-Hand, R-Shoulder                             & L-Wrist, L-Shoulder, Torso via R-Hand                              \\ \hline
1 & \textit{I}         & \textit{I}(2.9), \textit{I}(3.0), \textit{my}(5.0)             & \textit{I}(4.3), \textit{I}(4.4), \textit{my}(7.3)                \\
2 & \textit{deposit}   & \textit{to}(3.3), \textit{deposit}(4.1), \textit{deposit}(6.7) & \textit{to}(4.5), \textit{deposit}(6.8), \textit{deposit}(9.4) \\
3 & \textit{here}      & \textit{want}(3.4), \textit{now}(3.6), \textit{here}(3.9)      & \textit{here}(5.0), \textit{want}(5.1), \textit{do}(5.8)         \\ \hline
                                                                                                  \\ \hline
~ & Output & \multicolumn{2}{c}{Label weights}                                   \\
  & Label     & R-Hand, L-Hand, R-Shoulder                             & L-Wrist, L-Shoulder, Torso via R-Hand                              \\ \hline
1 & {\green \textit{I}}         & \textbf{\textit{I}(1.97)}, \textit{my}(0.58)                    & {\bf \textit{I}(1.33)}, \textit{my}(0.40)                      \\
2 & {\green \textit{deposit}}   & \textbf{\textit{deposit}(1.29)}, \textit{to}(1.00)              & {\bf \textit{deposit}(1.14)}, \textit{to}(1.00)                \\
3 & {\red \textit{want}}      & \textbf{\textit{want}(1.00)}, \textit{now}(0.94), \textit{here}(0.87)   & {\bf \textit{here}(0.68)}, \textit{want}(0.60), \textit{do}(0.59)      \\ \hline
\end{tabular}
}
% \todo[inline]{Make less wordy!}
\caption{KinTrans $k$NN-m queries showing how the predicted label is computed ($k=3$, $l(T)=2$, DTW).
The top shows true labels, $k$-closest sequences and their distances, for both sub-skeleton features.
The bottom shows weights for the labels (max sum in bold) and the resulting output label.
}
	\label{tab:interpretability_eg}
\end{table}

%% file: tables/datasets.tex
\begin{table*}[t] 
	\centering
{ \setlength{\tabcolsep}{0.5em}
	\begin{tabular}{l|lc|rrr|ccr}
		\hline
		Benchmark & Type & Jts. & Seq & Class & Sub & Frames/Seq & F./s & Seq/Cl. \\% & Ex. \\
		\hline
        KinTrans  & SL-Body   & $10$ & $5$,$166$ 	& $73$ 	& n/a  & $40 \pm 13$           & $30$  & $71$ \\% & \textit{how} \\
		% ASL  	  & SL-Body   & $18$ & $2$,$885$ 	& $100$	& n/a  & $89 \pm 39$           & $30$  & $29$ \\% & \textit{how} \\
        LM        & SL-Hands  & $54$ & $17$,$312$ 	& $60$ 	& $25$ & $71 \pm 21$           & $30$  & $289$ \\% & \textit{dog} \\
        NTU60     & HA-Body   & $25$ & $44$,$887$   & $60$  & $40$ & $78 \pm 34$           & $30$  & $748$ \\%& \textit{drop}\\
        UCF       & HA-Body   & $15$ & $1$,$280$ 	& $16$ 	& $16$ & $66 \pm 34$           & $30$  & $80$  \\%& \textit{hop} \\
        MHAD      & HA-Body   & $15$ & $660$ 	    & $11$ 	& $12$ & $3$,$602 \pm 2$,$510$ & $480$ & $60$  \\%& \textit{clap} \\
        \hline
	\end{tabular}
}

	\caption{Test data sets, showing type (Sign Language or Human Action), number of skeletal joints, sequences, classes and subjects; Mean$\pm$SD of frames per sequence, frames per second in Hz, and mean sequences per class.}
	\label{tab:datasets}
\end{table*}

% these are results with simplified datasets:

% 	\begin{tabular}{l|lc|rrrrrrl}
% 		\hline
% 		Benchmark & Type & Jts. & Seq & Class & Sub & Rep/Sub & Fr/Seq & Seq/Class & Examples \\
% 		\hline
%         KinTrans & SL-Body   & $10$ 	& $5$,$166$ 	& $73$ 	& n/a & n/a & $40.4$ & $70.8$ & \textit{America, bank, how, no} \\
%         LM      & SL-Hands	        & $54$ 	& $17$,$312$ 	& $60$ 	& $25$ & $11.5$ & $30.5$ & $288.5$ & \textit{brush, candy, dog, milk} \\
%         UCF & HA-Body         & $15$ 	& $1$,$280$ 	& $16$ 	& $16$ & $5.0$ & $35.0$ & $80.0$ & \textit{climb up, hop, kick, punch} \\
%         MHAD & HA-Body        & $15$ 	& $660$ 	& $11$ 	& $12$ & $5.0$ & $51.4$ & $60.0$ & \textit{clap hands, sit down} \\
%         \hline
% 	\end{tabular}

%% file: figures/LM_cut_eg.tex
\begin{figure}[b] \centering
    \includegraphics[width=\columnwidth, clip, trim=0 0 0 50]{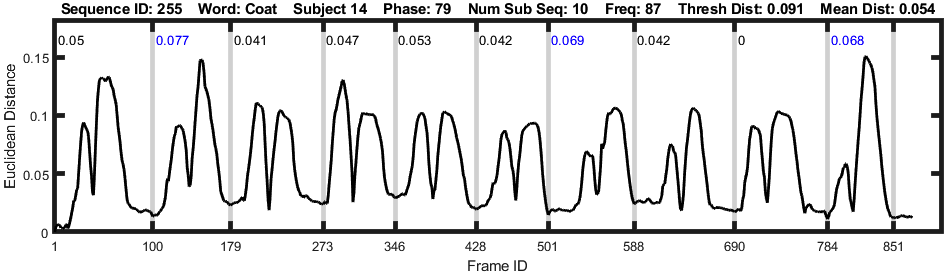}
    \\ \vspace{-.3cm}
    \caption{
    Example LM raw sequence for the sign \textit{coat} which is performed multiple times. 
    The Fourier transform identifies $10$ similar signs for \textit{coat} which are cut into individual segment sequences.
    % A centroid segment sequence is identified (minimum total distance to all other segment sequences), and distances are shown to the centroid (it has a distance of 0 to itself).
    }   \label{fig:LM_cut_eg}
\end{figure}

%% file: tables/feature_mining_results.tex
\begin{table}[]
  \centering
{
\begin{tabular}{l|r|l|ccl}
\hline
\multicolumn{3}{c|}{Feature Mining Input}                      & \multicolumn{3}{c}{Mining Results}                                                                                                   \\ 
Benchm.   & $|C|$                  & Method                      & \multicolumn{1}{r|}{$|T|$}                & \multicolumn{1}{r|}{$l(T)$}               & Sub-skeleton Features                                    \\ \hline
\multirow{4}{*}{\makecell[cl]{KinTrans \\ 20\% T/C}}   & \multirow{4}{*}{100} & 2NN-m(DTW)                 & \multicolumn{1}{r|}{5}                  & \multicolumn{1}{r|}{1}                  & abs(Rhand,Lwrist,Lelbow,neck,torso)                   \\ \cline{3-6} 
          &                      & DM-m(CF)                   & \multicolumn{1}{r|}{3}                  & \multicolumn{1}{r|}{1}                  & abs(Rhand,Lhand,Rshoulder)                            \\ \cline{3-6} 
          &                      & \multirow{2}{*}{DM-m(DTW)} & \multicolumn{1}{r|}{\multirow{2}{*}{6}} & \multicolumn{1}{r|}{\multirow{2}{*}{4}} & abs(Lelbow,Rwrist), Rhand(Relbow,Lelbow),           \\
          &                      &                             & \multicolumn{1}{r|}{}                   & \multicolumn{1}{r|}{}                   & Lshoulder(torso), Lhand(neck)                          \\ \hline
% \multirow{2}{*}{\makecell[cl]{ASL25\\ 60\% T/C}}  & \multirow{2}{*}{100}  & \multirow{2}{*}{DM-s(DTW)} & \multicolumn{1}{r|}{\multirow{2}{*}{5}} & \multicolumn{1}{r|}{\multirow{2}{*}{1}} & Rwrist(Lwrist,Rshoulder), Head(Lear),   \\ 
% &                      &                             & \multicolumn{1}{r|}{}                   & \multicolumn{1}{r|}{}                   & Neck(Rwrist,Reye)               \\ \cline{3-6}\hline
% \multirow{1}{*}{\makecell[cl]{ASL100}}  & \multirow{1}{*}{100}  & \multirow{1}{*}{DM-m(DTW)} & \multicolumn{1}{r|}{\multirow{1}{*}{4}} & \multicolumn{1}{r|}{\multirow{1}{*}{1}} & Rwrist(Lwrist,Reye,Relbow,Rshoulder)   \\ \hline
\multirow{5}{*}{\makecell[cl]{LM 5-fold \\ XSub}}  & \multirow{5}{*}{70}  & \multirow{2}{*}{1NN-s(CF)} & \multicolumn{1}{r|}{\multirow{2}{*}{8}} & \multicolumn{1}{r|}{\multirow{2}{*}{4}} & Rpalm(Rpinky,Rmid,Rthumb,Rring),   \\
          &                      &                             & \multicolumn{1}{r|}{}                   & \multicolumn{1}{r|}{}                   & Ridx(Rmid,Rthumb),Lidx(Lring),abs(Lpinky)               \\ \cline{3-6} 
          &                      & 3NN-m(CF)                  & \multicolumn{1}{r|}{5}                  & \multicolumn{1}{r|}{1}                  & Ridx(Rmid,Larm,Rpinky,Rring,Rarm)             \\ \cline{3-6} 
          &                      & \multirow{2}{*}{DM-m(DTW)} & \multicolumn{1}{r|}{\multirow{2}{*}{8}} & \multicolumn{1}{r|}{\multirow{2}{*}{4}} & Rpalm(Rring,Rpinky,Rmiddle,Rthumb),                 \\
          &                      &                             & \multicolumn{1}{r|}{}                   & \multicolumn{1}{r|}{}                   & abs(Lpinky),Ridx(Rmid,Rthumb),Lidx(Lring) \\ \hline
\end{tabular}
}
\caption{Sign language feature mining input and results. 
Input columns show the benchmark data set, number of canonical sub-skeletons, and classification method.
Result columns show number of chosen canonical sub-skeletons, number of features, and joint details (joints in brackets are relative to the joint outside of brackets and abs denotes absolute joints).
}
	\label{tab:feature_mining_results}
\end{table}

%% file: tables/accuracy_sign_lang.tex
\begin{table}[t!]
\begin{subfigure}[b]{.5\linewidth}   % \begin{minipage}[t]{.5\linewidth}
\centering
\begin{tabular}{l|llr}  
\hline
Benchmark                                       & Paper                 & Method      & Acc.       \\ \hline
\multirow{4}{*}{\makecell[cl]{KinTrans \\ 2 T/C}}    & \cite{yan2018}       & ST-GCN      & 57.6           \\ \cline{2-4} 
                                                & \multirow{3}{*}{Ours} & 2NN-m (DTW) & 79.4           \\
                                                &                       & DM-m (CF)   & 75.0           \\
                                                &                       & DM-m (DTW)  & \textbf{80.1}  \\ \hline
\multirow{4}{*}{\makecell[cl]{KinTrans \\ 3 T/C}}    & \cite{yan2018}       & ST-GCN      & 76.7           \\ \cline{2-4} 
                                                & \multirow{3}{*}{Ours} & 2NN-m (DTW) & 84.5           \\
                                                &                       & DM-m (CF)   & 84.9           \\
                                                &                       & DM-m (DTW)  & \textbf{90.2}  \\ \hline
% \multirow{4}{*}{\makecell[cl]{KinTrans \\ 5\% T/C}}  & \cite{yan2018}       & ST-GCN      & 81.2           \\ \cline{2-4} 
%                                                 & \multirow{3}{*}{Ours} & 2NN-m (DTW) & 89.2           \\
%                                                 &                       & DM-m (CF)   & 84.1           \\
%                                                 &                       & DM-m (DTW)  & \textbf{92.3}  \\ \hline
\multirow{4}{*}{\makecell[cl]{KinTrans \\ 10\% T/C}} & \cite{yan2018}       & ST-GCN      & 96.7           \\ \cline{2-4} 
                                                & \multirow{3}{*}{Ours} & 2NN-m (DTW) & 96.4           \\
                                                &                       & DM-m (CF)   & 96.2           \\
                                                &                       & DM-m (DTW)  & \textbf{98.4}  \\ \hline
\multirow{4}{*}{\makecell[cl]{KinTrans \\ 20\% T/C}} & \cite{yan2018}       & ST-GCN      & 99.2           \\ \cline{2-4} 
                                                & \multirow{3}{*}{Ours} & 2NN-m (DTW) & 99.3           \\
                                                &                       & DM-m (CF)   & 96.7           \\
                                                &                       & DM-m (DTW)  & \textbf{99.5} \\ \hline
% \multirow{2}{*}{\makecell[cl]{ASL25 \\  60\% T/C}}  & \cite{yan2018}       & ST-GCN      & 47.3           \\ \cline{2-4} 
%                                                 & \multirow{1}{*}{Ours} & DM-s (DTW) & \textbf{54.5}  \\ \hline
% \multirow{2}{*}{\makecell[cl]{ASL100 \\  60\% T/C}}  & \cite{yan2018}       & ST-GCN      & 27.3           \\ \cline{2-4} 
%                                                 & \multirow{1}{*}{Ours} & DM-m (DTW) & \textbf{29.1}  \\ \hline
\multirow{4}{*}{\makecell[cl]{LM \\ 5-fold XSub}}               & \cite{hernandez2020} & Kine.-LSTM  & \textit{91.1}  \\ \cline{2-4} 
                                                & \multirow{3}{*}{Ours} & 1NN-s (CF) & 80.6           \\
                                                &                       & 3NN-m (CF) & 71.3            \\
                                                &                       & DM-m (DTW)  & \textbf{94.3}  \\ \hline
\end{tabular}
\caption{%\tiny
Sign language data sets.
}
\label{tab:accuracy_sign_lang}
\end{subfigure}%
\begin{subfigure}[b]{.5\linewidth}    % \begin{minipage}[t]{.5\linewidth}
\centering
\begin{tabular}{l|llr}
\hline
Benchmark                      & Paper                        & Method                   & Acc.        \\ \hline
\multirow{2}{*}{\makecell[cl]{NTU60 \\ XSub}}  & \cite{yan2018}              & ST-GCN                   & \textbf{78.7}   \\ \cline{2-4} 
                              & Ours                         & DM-m (DTW)                    & $70.8$          \\ \hline
\multirow{2}{*}{\makecell[cl]{NTU60 \\ XView}} & \cite{yan2018}              & ST-GCN                   & \textbf{86.8}   \\ \cline{2-4} 
                              & Ours                         & DM-m (DTW)                    & $82.8$          \\ \hline
\multirow{10}{*}{\makecell[cl]{UCF \\ 4-fold}} & \cite{ellis2013}            & Log. Reg.                & \textit{95.9}          \\
                              & \cite{ohn2013joint}         & SVM                      & \textit{97.1}          \\
                              & \cite{presti2015hankelet}   & dHMM                     & \textit{97.7}          \\
                              & \cite{slama2015accurate}    & SVM                      & \textit{97.9}          \\
                              & \cite{zanfir2013moving}     & $k$NN+Vote             & \textit{98.5}          \\
                              & \cite{kerola2014spectral}   & SVM                      & \textit{98.8}          \\
                              & \cite{devanne2014}          & DP+$k$NN                 & \textit{99.2}          \\
                              & \cite{yan2018}              & ST-GCN                   & \textbf{99.7}   \\ \cline{2-4} 
                              & \multirow{2}{*}{Ours}        & 1NN-s (DTW)                    & $96.9$          \\
                              &                              & DM-m (DTW)                    & $99.5$          \\ \hline
\multirow{7}{*}{\makecell[cl]{MHAD \\ XSub}}   & \cite{ofli2013}             & $k$-SVM                  & \textit{80.0}          \\
                              & \cite{yan2018}              & ST-GCN                   & $89.8$          \\
                              & \cite{ofli2014sequence}     & SVM                      & \textit{95.4}          \\
                              & \cite{ijjina2014human}      & CNN                      & \textit{98.4}          \\
                              & \cite{chaudhry2013}         & MKL-SCM                  & \textit{\textbf{100}}    \\ \cline{2-4} 
                              & \multirow{2}{*}{Ours}        & 1NN-s (CF)                    & $94.9$          \\
                              &                              & DM-m (DTW)                    & \textbf{100}    \\ \hline
\end{tabular}
\caption{  %\tiny
Human Action data sets. 
}
\label{tab:accuracy_human_action}
\end{subfigure} 
\\ \vspace{-.6cm}
\caption{Classification accuracy results comparing our methods against others.
For sign language data sets, we achieve the best results by a large margin on low training information, our simple $k$NN-m classifier is often better than others (which use complex neural-networks), and our DM-m method performs best in all tests.
Although our focus is sign language recognition, our human action data set results show that our methods generalize well and achieve high accuracies. 
Accuracies in \textit{italics} were reported in previous work.
%, and \cite{hernandez2020} uses sequences from a non-publicly available manual segmentation process. %\footnote{Sequences are generated from a non-publicly available manual segmentation process.}
}
\end{table}

%% file: figures/testtime_small.tex
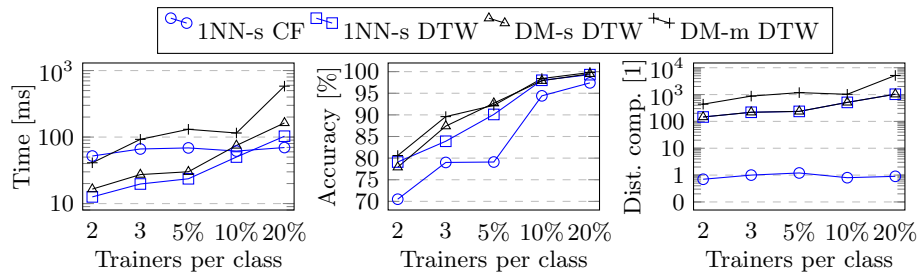
\begin{figure}[t]
\vspace{-.2cm}
\begin{tikzpicture}    %[scale=.65]
\pgfplotsset{every axis title/.append style={at={(0.5,0.85)}}}
\begin{groupplot}[
     group style = {group size = 3 by 1,
                    horizontal sep=1.25cm,
                    vertical sep=0.15cm,},
     width= .36\columnwidth, %5.5cm,
     height= .29\columnwidth, %3.6cm,
     grid style=dashed,
     xmode = normal,
     xmin=0.8, xmax=5.2,
     xtick style={draw=none},
     xtick=data,
     xticklabels={,,},
     ybar,
     ymajorgrids=true,
     ymode = log,
     log ticks with fixed point,
     log origin = infty,
    ]

\nextgroupplot[
sharp plot,
stack plots = false,
% legend pos= south east, %outer north east,
xticklabels={$2$, $3$, $5\%$, $10\%$, $20\%$},
xlabel={Trainers per class},
ymin=8, ymax=1300,
ytick={1,10,100,1000,10000},
yticklabels={$1$,$10$,$100$,$10^{3}$},
ylabel={Time [ms]},
ylabel shift = -0.20cm,
xticklabel shift = -0.10cm,
xlabel shift = -0.10cm,
y tick label style={font=\small},
% legend pos= south east, %outer north east,
legend style={
    at={(3.55,1.38)},  %(3.890,1.38)
    font=\small
    },
% height=.5\columnwidth,%3.6cm,
legend columns=4,
]

\addplot[
color=blue,
mark=o,
]
coordinates {(1,51.9)(2,66.2)(3,68.7)(4,61.6)(5,69.6)};

% \addplot[
% color=blue,
% mark=x,
% ]
% coordinates {(1,25.3)(2,32.76)(3,29.6)(4,32.9)(5,34.5)};

\addplot[
color=blue,
mark=square,
]
coordinates {(1,12.6)(2,19.7)(3,23.6)(4,50.1)(5,103.0)};

\addplot[
color=black,
mark=triangle,
]
coordinates {(1,16.5)(2,27.3)(3,30.1)(4,74.8)(5,163.2)};

\addplot[
color=black,
mark=+,
]
coordinates {(1,41.1)(2,92.4)(3,130.8)(4,115.1)(5,582.9)};

\legend{1NN-s CF}
% \addlegendentry{1NN-s CF-apx}
\addlegendentry{1NN-s DTW}
\addlegendentry{DM-s DTW}
\addlegendentry{DM-m DTW}

% \nextgroupplot[group/empty plot
% ]

\nextgroupplot[
sharp plot,
stack plots = false,
xticklabels={$2$, $3$, $5\%$, $10\%$, $20\%$},
xlabel={Trainers per class},
ymode = normal,
ymin=68, ymax=102,
ytick={60,70,75,80,85,90,95,100},
yticklabels={$60$,$70$,$75$,$80$,$85$,$90$,$95$,$100$},
ylabel={Accuracy~[\%]},
ylabel shift = -0.20cm,
x tick label style={font=\small},
y tick label style={font=\small},
xticklabel shift = -0.10cm,
xlabel shift = -0.10cm,
]

\addplot[
color=blue,
mark=o,
]
coordinates {(1,70.5)(2,79.0)(3,79.1)(4,94.4)(5,97.4)};

% \addplot[
% color=blue,
% mark=x,
% ]
% coordinates {(1,69.9)(2,74.2)(3,75.2)(4,93.1)(5,96.5)};

\addplot[
color=blue,
mark=square,
]
coordinates {(1,79.1)(2,83.9)(3,90.1)(4,98.0)(5,99.3)};

\addplot[
color=black,
mark=triangle,
]
coordinates {(1,77.9)(2,87.4)(3,92.7)(4,97.9)(5,99.5)};

\addplot[
color=black,
mark=+,
]
coordinates {(1,80.6)(2,89.6)(3,92.2)(4,98.4)(5,99.8)};

\nextgroupplot[
sharp plot,
stack plots = false,
xticklabels={$2$, $3$, $5\%$, $10\%$, $20\%$},
xlabel={Trainers per class},
ymin=0.05, ymax=15000,
ytick={0.1,1,10,100,1000,10000,100000},
yticklabels={$0$,$1$,$10$,$100$,$10^{3}$,$10^{4}$},
ylabel={Dist. comp. [1]},
ylabel shift = -0.20cm,
x tick label style={font=\small},
y tick label style={font=\small},
xticklabel shift = -0.10cm,
xlabel shift = -0.10cm,
]

\addplot[
color=blue,
mark=o,
]
coordinates {(1,0.7)(2,1.0)(3,1.2)(4,0.8)(5,0.9)};

% \addplot[
% color=blue,
% mark=x,
% ]
% coordinates {(1,0.1)(2,0.1)(3,0.1)(4,0.1)(5,0.1)};

\addplot[
color=blue,
mark=square,
]
coordinates {(1,146)(2,219)(3,236)(4,511)(5,1028)};

\addplot[
color=black,
mark=triangle,
]
coordinates {(1,146)(2,219)(3,236)(4,511)(5,1028)};

\addplot[
color=black,
mark=+,
]
coordinates {(1,438)(2,876)(3,1180)(4,1022)(5,5140)};

\end{groupplot}

\end{tikzpicture}
\vspace{-.7cm}
\caption{Query latency of our classifiers for various training sets sizes using the KinTrans data set.
Shown are avg. time per query (left), overall classification acc. (middle), and avg. num. of trajectory distance computations per query (right).
}

\label{fig:testtime_small}
\end{figure}

%% file: sec-supplement.tex
%%%%%%%%%%%%%%%%%%%%%%%%%%%%%%%%%%%%%%%%%%%%%%%%%%%%%%%%%%%%%%%%%%%%%%%%%%%%%%%%
\section{Further Data Set Information}
%%%%%%%%%%%%%%%%%%%%%%%%%%%%%%%%%%%%%%%%%%%%%%%%%%%%%%%%%%%%%%%%%%%%%%%%%%%%%%%%

We describe additional KinTrans, LM, and NTU60 data set statistics.

% \clearpage
\subsection{KinTrans Data Set}

The KinTrans data set has $10$ joints $G$: neck, torso, right-shoulder, right-elbow, right-wrist, right-hand, left-shoulder, left-elbow, left-wrist, and left-hand (see Figure~\ref{fig:kintrans_seq_eg} for an example sequence).
The data set has varying numbers of sequences per class which can make it more difficult for some classifiers to achieve high accuracy, since some classes are over or underrepresented in the training set.  
The sequences per class mean is $71$, min $18$, max $200$, and standard deviation $44$.
Table~\ref{tab:kintrans_dataset} shows the number of sequences per class label.

\input{tables/kintrans_dataset}

\input{figures/kintrans_seq_eg}

\clearpage
\subsection{LM Data Set and Segmentation Results}
The LM data set has $27$ joints $G$ for the right hand and arm: $3$ thumb joints, $4$ index finger joints, $4$ middle finger joints, $4$ ring finger joints, $4$ pinky finger joints, $1$ hand joint, $1$ palm joint, $3$ wrist joints, and $3$ elbow joints.
The same joints exist for the left hand and arm, giving a total of $54$ joints. 
Figure~\ref{fig:lm_seq_eg} shows an example LM sequence.
    
\input{tables/lm_dataset}

\input{figures/lm_seq_eg}

As discussed in the main paper, we implemented a simple, automated segmentation process to cut LM sequences into smaller sequences that each perform a single sign once.
The sequences per class mean is $288$, min $242$, max $347$, and standard deviation $22$. 
Table~\ref{tab:lm_dataset} shows the derived number of sequences per class label from our segmentation process.

\clearpage
\subsection{NTU60 Data Set of Single Subject Actions}
The NTU60 data set has $25$ joints $G$: hip, spine, neck, head, shoulder, right-shoulder, right-elbow, right-wrist, right-hand, right-fingertip, right-thumb, right-hip, right-knee, right-ankle, right-foot, left-shoulder, left-elbow, left-wrist, left-hand, left-fingertip, left-thumb, left-hip, left-knee, left-ankle, and left-foot.
Figure~\ref{fig:ntu60_seq_eg} shows an example NTU60 sequence.

Per the main paper, we extract only single-subject sequences from the NTU RGB+D 60-class human action data set, and insert them into a data set called NTU60.
The NTU60 sequences per class mean is $748$, min $16$, max $919$, and standard deviation $338$. 
Table~\ref{tab:ntu60_dataset} shows the derived number of sequences per class label.

\input{tables/ntu60_dataset}

\input{figures/ntu60_seq_eg}

\clearpage
%%%%%%%%%%%%%%%%%%%%%%%%%%%%%%%%%%%%%%%%%%%%%%%%%%%%%%%%%%%%%%%%%%%%%%%%%%%%%%%%
\section{Details of the Experimental Setup}
%%%%%%%%%%%%%%%%%%%%%%%%%%%%%%%%%%%%%%%%%%%%%%%%%%%%%%%%%%%%%%%%%%%%%%%%%%%%%%%%

We describe skeletal sequence normalization methods (translation, rotation, standardized limb lengths), and ST-GCN experiment parameter details. Further details can be found in our source code repository.

\subsection{Normalization Methods}

We briefly discuss the other three sequence normalization methods that our automated mining is allowed to choose for a data set.

A subject’s joints within a sequence can be \textit{translated} to the origin such that a key joint (e.g. neck) is at the origin for the first frame (or every frame if one chooses).

A subject’s joints can be \textit{rotated} about the x, y, or z axis relative to the first frame or every frame via rotation matrices and matrix multiplication. 
Examples include rotating the subject to face the camera with their hip directly beneath the neck or having the back of the hands face the camera with the wrist directly beneath the knuckles. 
This can be toggled such that, for example, the subject faces the camera for the first frame or every frame. 
The rotation and translation normalizations can be useful when various subjects start from disparate coordinates or have different azimuth and elevations.

A subject’s limbs can be normalized to \textit{standardized lengths} using linear algebra: for each frame start at a center skeletal joint and work towards the extremities, modifying limb vector magnitudes to the standard length, while preserving the original limb vector direction. 
Standard limb lengths are pre-computed by analyzing subjects in a data set and obtaining mean (standard) limb lengths. This normalization can help when there are substantial body size differences between subjects.

\subsection{ST-GCN Experiment Parameters}

The following parameters were used to obtain the best accuracy in ST-GCN experiments on the high-performance computing environment: 
%$1$-CPU, $1$-GPU, $20$ GB RAM,
batch\_size: $20$, epochs: $80$, base learning rate: $0.1$, weight\_decay: $0.0001$, optimizer: SGD, nesterov: True.
The ST-GCN training runtimes on the KinTrans data set for the $2$, $3$, $5\%$, $10\%$, and $20\%$ trainers per class experiments were $106$, $156$, $170$, $358$, and $713$ minutes, respectively (times shown are the fastest of three independent runs for each experiment).

\clearpage

%%%%%%%%%%%%%%%%%%%%%%%%%%%%%%%%%%%%%%%%%%%%%%%%%%%%%%%%%%%%%%%%%%%%%%%%%%%%%%%%
\section{Additional Results}
%%%%%%%%%%%%%%%%%%%%%%%%%%%%%%%%%%%%%%%%%%%%%%%%%%%%%%%%%%%%%%%%%%%%%%%%%%%%%%%%

We show additional experimental results on: (i) classifier interpretability, (ii) detailed KinTrans accuracies, (iii) DM scalability, and (iv) feature mining. %, and (v) a mining discovery that identifies a surprisingly information-rich joint.

\subsection{Classifier Interpretability}

\begin{figure}[p]
    \centering
    \includegraphics[width=.7\columnwidth]{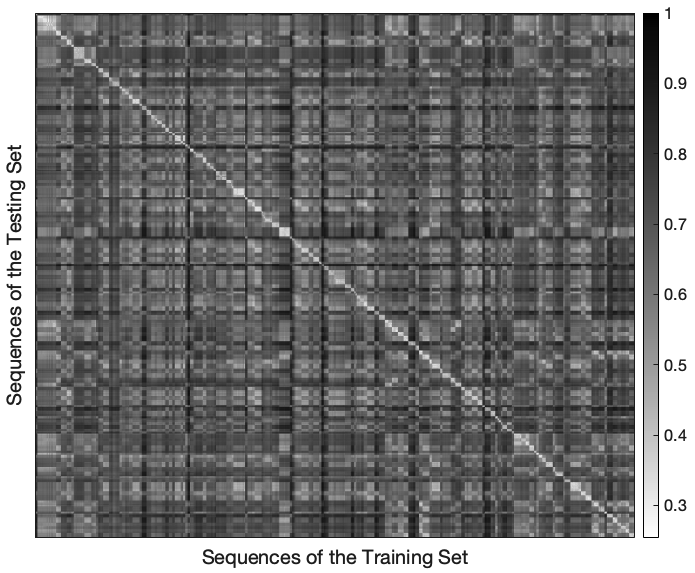}
    \caption{
    The KinTrans '1NN-s DTW' $10\%$ trainers per class heat map, which contains the distances (normalized to $[0,1]$) of derived feature trajectories.
    There are $4,655$ testing data rows and $511$ training data columns, which are both sorted by $73$ class labels.
    }   \label{fig:nn_heat_map}
    \vspace{-.2cm}
\end{figure}

\begin{figure}[p]
    \centering
    \includegraphics[width=.7\columnwidth]{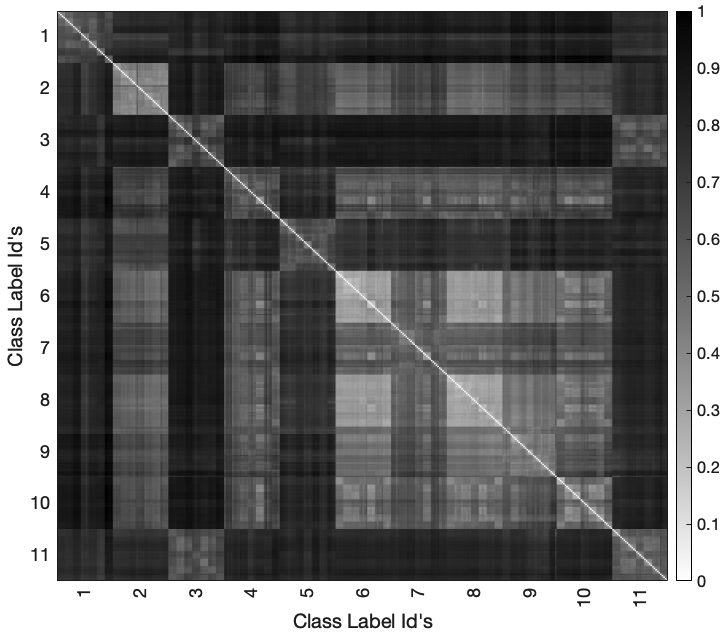}
    \caption{
    Training distance matrix heat map for the MHAD DM-m experiment.
    The DM contains derived feature trajectory distances normalized to $[0,1]$, has size $384 \times 384$, and rows/columns are sorted by class label then subject. 
    % Class Label Id's: $1$ - \textit{bending}, $2$ - \textit{clapping hands}, $3$ - \textit{jumping jacks}, $4$ - \textit{jumping in place}, $5$ - \textit{punching}, $6$ - \textit{sit down}, $7$ - \textit{sit down stand up}, $8$ - \textit{stand up}, $9$ - \textit{throwing ball}, $10$ - \textit{waving right hand}, $11$ - \textit{waving two hands}.
    }   \label{fig:mhad_heat_map}
\end{figure}

The use of trajectory similarity measures as feature predictor inputs make it easy to interpret 
% how well a classifier will perform.
what a classifier predicted.
Closer distances signify greater similarity and further distances less similarity.
The greater the separation of distances between classes, the better the classifier will perform.
For example, if the KinTrans sequences for the class \textit{doctor} all have small distances to each other and large distances to other classes, then the classifier will be able to easily distinguish and correctly classify this sign.
% Other papers that use joint coordinates as feature input do not have this property, i.e. it is difficult to gain intuition on how well the classifier will perform based solely on joint coordinates.
Our feature template mining algorithm suggests a feature trajectory that will best separate classes.

We achieve classifier interpretability by visualizing trajectory distances in a heat map, where is it easy to compare distances amongst and between classes, thus providing intuition on the performance of the mined classifier.

%\clearpage
\subsubsection{Nearest Neighbor (NN)}
The heat map in Figure~\ref{fig:nn_heat_map} shows distances between the training and testing trajectories (using the 1NN-s trajectory feature template derived from the feature mining process).
The correct class labels are along the diagonal, and as the figure shows, the diagonal tends to be a lighter color (represents a closer distance) compared to areas outside the diagonal.
It is easy to determine which class is predicted for each test sequence simply by locating the smallest distance.

% \newpage
% \vspace{-0.35cm}
%\clearpage
\subsubsection{Distance Matrix (DM)}
Figure~\ref{fig:mhad_heat_map} helps one visualize why the distance matrix classifier achieves good classification accuracy results. 
The larger squares along the diagonal represent individual class labels that are compared to themselves, and each of these are typically a lighter color (represents a closer distance) since different sequences in the same class tend to be close to one another. Importantly, when one compares a larger square in the diagonal with its corresponding row, the other larger squares in the row tend to be a darker color (represents a further distance) since they are different classes. It is this ability to correctly separate close and far sequences that results in good classification accuracies. Underpinning this are feature templates and trajectory distance measures, which are used to compute the distance measures in the DM.

\clearpage
\subsection{Accuracy Comparison of Proposed Methods}
Detailed experiment results of our methods vs. ST-GCN for the KinTrans data set are in Table~\ref{tab:accuracy_kintrans}.
Our simpler $k$NN classifiers often perform close to or on-par with the more complex DM method.
Of particular note is our accuracy results with the simplest NN-s classifier and DTW, which beats the more complicated ST-GCN method in low-training information experiments.
\input{tables/accuracy_kintrans.tex}

\clearpage
\subsection{Scalability of DM Query Time }
Table~\ref{tab:dm_scale} shows the impact of reducing DM-m distance matrix feature columns on the KinTrans $20\%$ trainers per class protocol.
For DM-m(c) just one trainer per class is randomly selected for DM feature columns, and compared to DM-m(a) which uses all trainers per class for feature columns, classification accuracy only drops $0.1\%$ to $99.4\%$, but query runtimes are $9.1$ times faster.
Clearly, the DM method can scale to larger data sets by reducing feature columns, which can result in similar accuracy.
\input{tables/dm_scale}

\clearpage
\subsection{Feature Mining}

Supplementary feature mining results in Table~\ref{tab:feature_mining_results2} show that the greedy mining chooses a small number of discriminative sub-skeleton features across all sign language and human action benchmarks.
Though mined sub-skeleton features vary depending on the classifier method they typically show largely similar joint sets in the selected features.

\input{tables/feature_mining_results2.tex}

%% file: tables/kintrans_dataset.tex
\begin{table}[h]
\centering
\begin{tabular}{lr|lr|lr}
\hline
Label               & Num. & Label             & Num. & Label          & Num \\ 
\hline
\textit{America}          & 23   & \textit{great}          & 59   & \textit{please}      & 90  \\
\textit{I}                & 200  & \textit{have}           & 110  & \textit{request}     & 90  \\
\textit{ID}               & 90   & \textit{head}           & 20   & \textit{setup}       & 90  \\
\textit{allowed}          & 20   & \textit{help}           & 120  & \textit{sick}        & 20  \\
\textit{and}              & 190  & \textit{here}           & 155  & \textit{small bill}  & 90  \\
\hline
\textit{baggage}          & 20   & \textit{hi}             & 90   & \textit{stay}        & 20  \\
\textit{bank}             & 90   & \textit{high}           & 20   & \textit{system}      & 18  \\
\textit{bank statement}   & 90   & \textit{hotel}          & 20   & \textit{taxi}        & 20  \\
\textit{can}              & 20   & \textit{how}            & 120  & \textit{teller}      & 90  \\
\textit{check}            & 61   & \textit{how much}       & 90   & \textit{temperature} & 20  \\
\hline
\textit{cheque}           & 90   & \textit{hurts}          & 21   & \textit{thank you}   & 170 \\
\textit{claim}            & 20   & \textit{insurance card} & 20   & \textit{they}        & 20  \\
\textit{company account}  & 89   & \textit{interest rate}  & 90   & \textit{this}        & 20  \\
\textit{credit card}      & 90   & \textit{international}  & 90   & \textit{to}          & 30  \\
\textit{customer service} & 90   & \textit{manager}        & 90   & \textit{today}       & 90  \\
\hline
\textit{days}             & 20   & \textit{many}           & 19   & \textit{verfication} & 90  \\
\textit{deposit}          & 90   & \textit{medical}        & 61   & \textit{visit}       & 60  \\
\textit{diabetes}         & 20   & \textit{money}          & 90   & \textit{want}        & 90  \\
\textit{dizzy}            & 21   & \textit{monthly fee}    & 90   & \textit{what}        & 90  \\
\textit{do}               & 30   & \textit{my}             & 119  & \textit{where}       & 110 \\
\hline
\textit{doctor}           & 20   & \textit{need}           & 90   & \textit{with}        & 20  \\
\textit{email}            & 90   & \textit{new account}    & 90   & \textit{withdraw}    & 90  \\
\textit{family}           & 60   & \textit{no}             & 90   & \textit{yes}         & 150 \\
\textit{feel pain}        & 20   & \textit{now}            & 90   &             &     \\
\textit{fund transfer}    & 90   & \textit{pay balance}    & 90   &             &     \\
\hline
\end{tabular}
	\caption{KinTrans data set class labels and number of sequences per class. The classes are common words used in the airport, bank, and doctor's office.}
	\label{tab:kintrans_dataset}
\end{table}

%% file: figures/kintrans_seq_eg.tex
% \begin{figure}[H]\centering \vspace{-.3cm}  
\begin{figure}[b]\centering %\vspace{-.3cm}  

    \includegraphics[height=10cm]{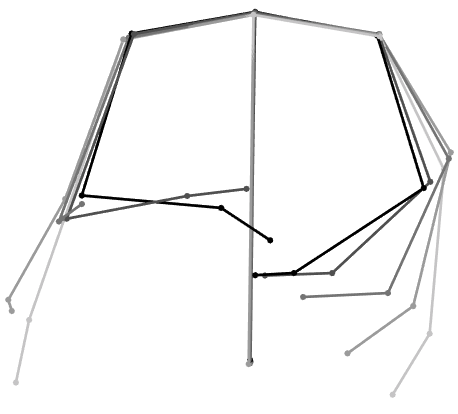}

\caption{
Kintrans sequence example showing joint and limb movement from earlier (lighter) to later (darker) time frames.}
	\label{fig:kintrans_seq_eg} 
%  	\vspace{-.3cm}
\end{figure}

%% file: tables/lm_dataset.tex
\begin{table}[h]
\centering

\begin{tabular}{lrr|lrr|lrr}
\hline
\multirow{2}{*}{Label} & Num.    & Num.   & \multirow{2}{*}{Label} & Num.    & Num.   & \multirow{2}{*}{Label} & Num.    & Num.   \\
                       & man. & auto &                        & man. & auto &                        & man. & auto \\
\hline
\textit{0}         & 285   & 309    & \textit{coat}   & 283  & 271    & \textit{more}   & 315  & 320                     \\
\textit{1}         & 322   & 319    & \textit{cold}   & 345  & 347    & \textit{orange} & 287  & 299                     \\
\textit{2}         & 308   & 307    & \textit{come}   & 296  & 306    & \textit{pig}    & 277  & 266                     \\
\textit{3}         & 277   & 291    & \textit{cost}   & 290  & 298    & \textit{please} & 262  & 276                     \\
\textit{4}         & 296   & 303    & \textit{cry}    & 280  & 281    & \textit{red}    & 272  & 283                     \\
\hline
\textit{5}         & 320   & 314    & \textit{dad}    & 278  & 285    & \textit{shoes}  & 311  & 315                     \\
\textit{6}         & 299   & 317    & \textit{deaf}   & 223  & 243    & \textit{small}  & 279  & 279                     \\
\textit{7}         & 283   & 298    & \textit{dog}    & 272  & 280    & \textit{socks}  & 265  & 270                     \\
\textit{8}         & 275   & 292    & \textit{drink}  & 280  & 279    & \textit{stop}   & 277  & 278                     \\
\textit{9}         & 297   & 298    & \textit{egg}    & 307  & 309    & \textit{store}  & 272  & 283                     \\
\hline
\textit{10}        & 260   & 242    & \textit{finish} & 323  & 326    & \textit{thanks} & 270  & 273                     \\
\textit{big}       & 251   & 277    & \textit{go}     & 293  & 319    & \textit{warm}   & 299  & 302                     \\
\textit{blue}      & 239   & 242    & \textit{good}   & 315  & 315    & \textit{water}  & 260  & 283                     \\
\textit{brush}     & 275   & 296    & \textit{green}  & 268  & 267    & \textit{what}   & 287  & 296                     \\
\textit{bug}       & 252   & 257    & \textit{happy}  & 275  & 274    & \textit{when}   & 263  & 266                     \\
\hline
\textit{candy}     & 258   & 282    & \textit{hot}    & 288  & 292    & \textit{where}  & 256  & 293                     \\
\textit{car-drive} & 250   & 264    & \textit{hungry} & 291  & 295    & \textit{why}    & 254  & 266                     \\
\textit{cat}       & 262   & 267    & \textit{hurt}   & 286  & 293    & \textit{with}   & 305  & 308                     \\
\textit{cereal}    & 250   & 264    & \textit{milk}   & 301  & 308    & \textit{work}   & 289  & 293                     \\
\textit{clothes}   & 292   & 292    & \textit{mom}    & 298  & 300    & \textit{yellow} & 247  & 244                     \\
\hline
\end{tabular}
	\caption{LM data set class labels and number of segmented sequences per class.
	Columns compare the manual segmentation results of Hernandez et al. that are not publicly available (total 16,890) vs. the automated  method in our implementation (total 17,312).
	There are $25$ sequences per class label prior to segmentation.}
	\label{tab:lm_dataset}
\end{table}

%% file: figures/lm_seq_eg.tex
% \begin{figure}[H]\centering \vspace{-.3cm}  
\begin{figure}[b]\centering %\vspace{-.3cm}  

\includegraphics[height=6.5cm]{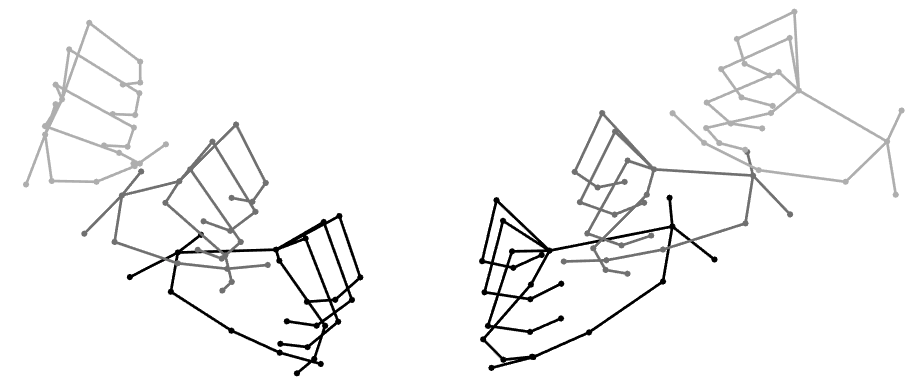}

\caption{
LM sequence example showing joint and limb movement from earlier (lighter) to later (darker) time frames.}
	\label{fig:lm_seq_eg} 
%  	\vspace{-.3cm}
\end{figure}

%% file: tables/ntu60_dataset.tex
\begin{table}[h]
\centering
\begin{tabular}{lr|lr}
\hline
Label                                & Num. & Label                                          & Num. \\
\hline
\textit{brushing hair}                     & 916  & \textit{reach into pocket}                           & 907  \\
\textit{brushing teeth}                    & 912  & \textit{reading}                                     & 907  \\
\textit{check time (from watch)}           & 906  & \textit{rub two hands together}                      & 908  \\
\textit{cheer up}                          & 915  & \textit{salute}                                      & 903  \\
\textit{clapping}                          & 907  & \textit{shake head}                                  & 903  \\
\hline
\textit{cross hands in front (say stop)}   & 891  & \textit{sitting down}                                & 909  \\
\textit{drink water}                       & 908  & \textit{sneeze-cough}                                & 902  \\
\textit{drop}                              & 914  & \textit{staggering}                                  & 896  \\
\textit{eat meal-snack}                    & 913  & \textit{standing up (from sitting position)}         & 919  \\
\textit{falling}                           & 894  & \textit{take off a hat-cap}                          & 915  \\
\hline
\textit{giving something to other person}  & 29   & \textit{take off a shoe}                             & 915  \\
\textit{hand waving}                       & 917  & \textit{take off glasses}                            & 913  \\
\textit{handshaking}                       & 20   & \textit{take off jacket}                             & 919  \\
\textit{hopping (one foot jumping)}        & 905  & \textit{taking a selfie}                             & 906  \\
\textit{hugging other person}              & 17   & \textit{tear up paper}                               & 904  \\
\hline
\textit{jump up}                           & 906  & \textit{throw}                                       & 916  \\
\textit{kicking other person}              & 35   & \textit{touch back (backache)}                       & 895  \\
\textit{kicking something}                 & 907  & \textit{touch chest (stomach-heart pain)}        & 894  \\
\textit{make a phone call-answer phone}    & 906  & \textit{touch head (headache)}                       & 894  \\
\textit{nausea or vomiting condition}      & 894  & \textit{touch neck (neckache)}                       & 895  \\
\hline
\textit{nod head/bow}                      & 903  & \textit{touch other persons pocket}                  & 28   \\
\textit{pat on back of other person}       & 20   & \textit{typing on a keyboard}                        & 912  \\
\textit{pickup}                            & 915  & \textit{use a fan (hand/paper)-feel warm} & 895  \\
\textit{playing with phone-tablet}         & 906  & \textit{walking apart from each other}               & 37   \\
\textit{point finger at the other person}  & 16   & \textit{walking towards each other}                  & 183  \\
\hline
\textit{pointing to something with finger} & 903  & \textit{wear a shoe}                                 & 913  \\
\textit{punching-slapping other person}    & 42   & \textit{wear jacket}                                 & 918  \\
\textit{pushing other person}              & 28   & \textit{wear on glasses}                             & 910  \\
\textit{put on a hat-cap}                  & 913  & \textit{wipe face}                                   & 904  \\
\textit{put the palms together}            & 903  & \textit{writing}                                     & 906  \\
\hline
\end{tabular}
	\caption{NTU60 data set class labels and number of sequences per class.}
	\label{tab:ntu60_dataset}
\end{table}

%% file: figures/ntu60_seq_eg.tex
% \begin{figure}[H]\centering \vspace{-.3cm}  
\begin{figure}[b]\centering %\vspace{-.3cm}  

    \includegraphics[height=12cm]{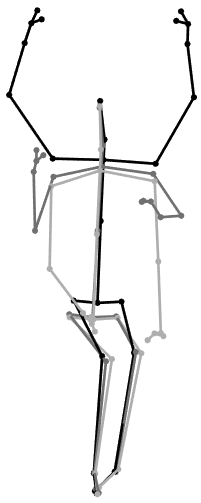}

\caption{
NTU60 sequence example showing joint and limb movement from earlier (lighter) to later (darker) time frames.}
	\label{fig:ntu60_seq_eg} 
%  	\vspace{-.3cm}
\end{figure}

%% file: tables/accuracy_kintrans.tex
\begin{table}[h]
\centering
{
\begin{tabular}{l|r|lrrrrrr}
\hline
Trainers %\footnote{4-trainers per class have similar results to 5\%-trainers per class and are omitted.}
& \multicolumn{1}{c|}{\multirow{2}{*}{ST-GCN}} & \multicolumn{7}{c}{Ours}                                             \\
per class             & \multicolumn{1}{c|}{}    & Dist. & NN-s          & NN-m   & 2NN-s   & 2NN-m   & DM-s    & DM-m           \\ \hline
\multirow{2}{*}{2}    & \multirow{2}{*}{$57.6$}  & CF    & $65.3$        & $66.3$ & $69.6$  & $66.3$  & $63.3$  & $75.0$         \\
                      &                          & DTW   & \textbf{80.2} & $79.4$ & $76.4$  & $79.4$  & $74.6$  & $80.1$         \\ \hline
\multirow{2}{*}{3}    & \multirow{2}{*}{$76.7$}  & CF    & $76.6$        & $77.2$ & $76.4$  & $77.2$  & $77.3$  & $84.9$         \\
                      &                          & DTW   & $84.6$        & $82.9$ & $84.5$  & $84.5$  & $90.0$  & \textbf{90.2}  \\ \hline
\multirow{2}{*}{5\%}  & \multirow{2}{*}{$81.2$}  & CF    & $79.9$        & $80.3$ & $81.7$  & $80.3$  & $79.3$  & $84.1$         \\
                      &                          & DTW   & $89.0$        & $89.2$ & $89.0$  & $89.2$  & $92.2$  & \textbf{92.3}  \\ \hline
\multirow{2}{*}{10\%} & \multirow{2}{*}{$96.7$}  & CF    & $85.5$        & $94.9$ & $94.3$  & $94.9$  & $91.7$  & $96.2$         \\
                      &                          & DTW   & $96.7$        & $96.4$ & $96.7$  & $96.4$  & $97.7$  & \textbf{98.4}  \\ \hline
\multirow{2}{*}{20\%} & \multirow{2}{*}{$99.2$}  & CF    & $91.8$        & $97.4$ & $97.4$  & $97.4$  & $96.7$  & $96.7$         \\
                      &                          & DTW   & $99.3$        & $99.3$ & $99.3$  & $99.3$  & $99.5$ & \textbf{99.5} \\ \hline
\end{tabular}
}

\caption{KinTrans American sign language data set classification accuracy results comparing our methods vs. ST-GCN on training sets of increasing sizes.}
	\label{tab:accuracy_kintrans}
\end{table}

% \begin{tabular}{l|r|rrr}
% \hline
% Trainers & \multirow{2}{*}{ST-GCN} & \multicolumn{3}{c}{Ours\footnote{$k$NN-s results are omitted as they did not outperform NN-s on the KinTrans data set.}}     \\
% per class &                         & NN-s    & DM-s       & DM-m           \\ \hline
% 2                 & 57.6                    & 79.1 & 77.9     & \textbf{80.6}          \\
% 3                 & 76.7                    & 83.9 & 87.4     & \textbf{89.6}          \\
% 5\%               & 81.2                    & 90.1          & \textbf{92.7}     & 92.2 \\
% 10\%              & 96.7                    & 98.0          & 97.9     & \textbf{98.4} \\
% 20\%              & 99.2                    & 99.3          & 99.5     & \textbf{99.8} \\

% \hline
% \end{tabular}

%% file: tables/dm_scale.tex
\begin{table}[h]
\centering
\begin{tabular}{llr|rrr}
\hline
Distance & DM Feature                & Num. Feature & Avg. Query & \multirow{2}{*}{Accuracy} & Avg. Distance \\
Matrix   & Col. Selection            & Columns      & Time (ms)  &                           & Computations  \\ \hline
DM-m(a)  & all $\mathbb{D}_r$/class  & $1$,$208$    & $273$      & $99.5$                    & $4$,$112$     \\
DM-m(b)  & 2 $\mathbb{D}_r$/class    & $146$        & $61$       & $99.6$                    & $584$         \\
DM-m(c)  & 1 $\mathbb{D}_r$/class    & $73$         & $30$       & $99.4$                    & $292$         \\ \hline
\end{tabular}
\caption{Impact on runtime and accuracy when reducing distance matrix feature columns with the KinTrans $20\%$ trainers per class protocol. Columns show DM feature column selection criteria, number of distance matrix feature columns, average query time, overall classification accuracy, and average number of trajectory distance computations per query. DM-m(a) contains all feature columns.  DM-m(b) and DM-m(c) randomly select $2$ or $1$ trainers per class from $\mathbb{D}_r$ for the feature columns, respectively.
}
\label{tab:dm_scale}
% \vspace{-0.3cm}
\end{table}

% \begin{tabular}{llr|rrr}
% \hline
% Distance  & & Feature & Query & \multicolumn{1}{c}{\multirow{2}{*}{Accuracy}} & \multicolumn{1}{c}{\multirow{2}{*}{Distance}} \\
% Matrix    & & columns & Time (ms) & \multicolumn{1}{c}{}                          & \multicolumn{1}{c}{}                          \\ \hline
% DM-m(a)    & & $1$,$028$    & $582$  & $99.8$                                          & $5$,$140$                                         \\
% DM-m(b) & & $146$     & $126$  & $99.8$                                          & $730$                                           \\
% DM-m(c) & & $73$      & $49$  & $99.5$                                          & $365$                                           \\ \hline
% \end{tabular}

%% file: tables/feature_mining_results2.tex
\begin{table}[h]
  \centering
{
\scalebox{0.85}{
\begin{tabular}{l|r|l|rrl}
\hline
\multicolumn{3}{c|}{Feature Mining Input}                     & \multicolumn{3}{c}{Mining Results}                                                                                                      \\
Benchm.  & $C$                    & Method                      & \multicolumn{1}{r|}{$T$}                   & \multicolumn{1}{r|}{$l(T)$}                 & Sub-skeleton Features                              \\ \hline
\multirow{5}{*}{\makecell[cl]{KinTrans \\ 5\% T/C}} & \multirow{5}{*}{100} & 1NN-s (DTW)                  & \multicolumn{1}{r|}{3}                   & \multicolumn{1}{r|}{2}                  & Rhand(Lwrist,Lhand), Lwrist(torso)                 \\ \cline{3-6} 
        &                      & 2NN-m (DTW)                  & \multicolumn{1}{r|}{3}                   & \multicolumn{1}{r|}{1}                  & Rhand(Lwrist,torso,Lhand)                          \\ \cline{3-6} 
         &                      & DM-m (CF)                   & \multicolumn{1}{r|}{3}                   & \multicolumn{1}{r|}{2}                  & Rwrist(Lhand,Lelbow), Rhand(torso)                 \\ \cline{3-6} 
         &                      & \multirow{2}{*}{DM-m (DTW)} & \multicolumn{1}{r|}{\multirow{2}{*}{5}}  & \multicolumn{1}{r|}{\multirow{2}{*}{3}} & Rhand(Lelbow,Lwrist,Rwrist),                       \\
         &                      &                             & \multicolumn{1}{r|}{}                    & \multicolumn{1}{r|}{}                   & Rwrist(neck), Lhand(Rhand)                         \\ \hline
\multirow{6}{*}{\makecell[cl]{KinTrans \\ 10\% T/C}} & \multirow{6}{*}{100} & 1NN-s (DTW)                  & \multicolumn{1}{r|}{4}                   & \multicolumn{1}{r|}{3}                  & Rhand(torso,Lhand), Lshoulder(torso), torso(Rhand) \\ \cline{3-6} 
        &                      & 2NN-m (DTW)                 & \multicolumn{1}{r|}{3}                   & \multicolumn{1}{r|}{1}                  & Rhand(torso,Lhand,Lshoulder)                            \\ \cline{3-6} 
         &                      & \multirow{2}{*}{DM-m (CF)}  & \multicolumn{1}{r|}{\multirow{2}{*}{6}}  & \multicolumn{1}{r|}{\multirow{2}{*}{6}} & Rhand(neck),   Relbow(Lhand), abs(Rhand),          \\
         &                      &                             & \multicolumn{1}{r|}{}                    & \multicolumn{1}{r|}{}                   & Rwrist(Relbow), Lelbow(Lshoulder), neck(Rhand)     \\ \cline{3-6} 
         &                      & \multirow{2}{*}{DM-m (DTW)} & \multicolumn{1}{r|}{\multirow{2}{*}{6}}  & \multicolumn{1}{r|}{\multirow{2}{*}{4}} & Rhand(Lelbow,Lwrist,Rwrist), neck(Rhand),          \\
         &                      &                             & \multicolumn{1}{r|}{}                    & \multicolumn{1}{r|}{}                   & Relbow(Rhand), Lwrist(Relbow)                      \\ \hline
\multirow{3}{*}{\makecell[cl]{NTU60 \\ XSub}}    & \multirow{3}{*}{90}  & \multirow{3}{*}{DM-m (DTW)} & \multicolumn{1}{r|}{\multirow{3}{*}{10}} & \multicolumn{1}{r|}{\multirow{3}{*}{5}} & abs(Rhand,neck,spine,head,Relbow,shoulder),        \\
           &                      &                             & \multicolumn{1}{r|}{}                    & \multicolumn{1}{r|}{}                   & spine(hip), neck(shoulder),                        \\
         &                      &                             & \multicolumn{1}{r|}{}                    & \multicolumn{1}{r|}{}                   & Lhand(Lfingertip), R-hand(Rfingertip)              \\ \hline
\multirow{3}{*}{\makecell[cl]{NTU60 \\ XView}}    & \multirow{3}{*}{90}  & \multirow{3}{*}{DM-m (DTW)} & \multicolumn{1}{r|}{\multirow{3}{*}{10}} & \multicolumn{1}{r|}{\multirow{3}{*}{5}} & abs(Rwrist,Lwrist,shoulder,spine,Lhand),           \\
         &                      &                             & \multicolumn{1}{r|}{}                    & \multicolumn{1}{r|}{}                   & hip(Rfingertip,Rhip), spine(hip),                  \\
         &                      &                             & \multicolumn{1}{r|}{}                    & \multicolumn{1}{r|}{}                   & head(Lhand), Relbow(Rwrist)                        \\ \hline
\multirow{3}{*}{\makecell[cl]{UCF \\ 4-fold}}      & \multirow{3}{*}{225} & 1NN-s (DTW)                  & \multicolumn{1}{r|}{4}                   & \multicolumn{1}{r|}{3}                  & abs(Relbow,torso), Rhand(Lelbow), Rknee(torso)     \\ \cline{3-6} 
         &                      & \multirow{2}{*}{DM-m (DTW)} & \multicolumn{1}{r|}{\multirow{2}{*}{6}}  & \multicolumn{1}{r|}{\multirow{2}{*}{5}} & abs(Lelbow,head), neck(Relbow), torso(Relbow),     \\
         &                      &                             & \multicolumn{1}{r|}{}                    & \multicolumn{1}{r|}{}                   & Relbow(Lelbow), Lfoot(Lelbow)                      \\ \hline
\multirow{3}{*}{\makecell[cl]{MHAD \\ XSub}}     & \multirow{3}{*}{225} & \multirow{2}{*}{1NN-s (CF)} & \multicolumn{1}{r|}{\multirow{2}{*}{5}}  & \multicolumn{1}{r|}{\multirow{2}{*}{4}} & Lfoot(Relbow), Lelbow(Lshoulder),                  \\
         &                      &                             & \multicolumn{1}{r|}{}                    & \multicolumn{1}{r|}{}                   & Relbow(head,Lelbow), Lhand(torso)                  \\ \cline{3-6} 
         &                      & DM-m (DTW)                  & \multicolumn{1}{r|}{4}                   & \multicolumn{1}{r|}{3}                  & Rfoot(Lhand,head), torso(head), Lshoulder(neck)    \\ \hline
\end{tabular}
}
}
\caption{Additional sign language and human action feature mining input and results. 
Input columns show the benchmark data set, number of canonical sub-skeletons, and classification method.
Result columns show number of chosen canonical sub-skeletons, number of features, and joint details (joints in brackets are relative to the joint outside of brackets and abs denotes absolute joints).
}
	\label{tab:feature_mining_results2}
\end{table}